%% file: main.tex
\documentclass[10pt,twocolumn,letterpaper]{article}

\usepackage{cvpr}              
\input{preamble}
\definecolor{cvprblue}{rgb}{0.21,0.49,0.74}
\usepackage[pagebackref,breaklinks,colorlinks,allcolors=cvprblue]{hyperref}
\usepackage{graphicx}
\usepackage{subcaption}
\usepackage{multirow}
\usepackage{amsmath}
\usepackage{paralist}
\usepackage{wrapfig}

\def\paperID{18659} 
\def\confName{CVPR}
\def\confYear{2026}

\title{Can Natural Image Autoencoders Compactly Tokenize fMRI Volumes \\for Long-Range Dynamics Modeling?}

\author{Peter Yongho Kim$^{1}$\thanks{Equal contribution.} , Juhyeon Park$^{2}$\footnotemark[1] , Jungwoo Park$^{1}$, \\Jubin Choi$^{2}$, Jungwoo Seo$^{3}$, Jiook Cha$^{2,4}$, Taesup Moon$^{1,2,5}$\thanks{Corresponding author.} \\
$^{1}$ECE, Seoul National University,
$^{2}$IPAI, Seoul National University, \\
$^{3}$BCS, Seoul National University,
$^{4}$Psychology, Seoul National University, \\
$^{5}$ASRI / INMC / AIIS, Seoul National University\\
\texttt{\{peterkim98, parkjh9229, tsmoon\}@snu.ac.kr} \\
}
\newcommand{\ours}{TABLeT}
\newcommand{\meanstd}[2]{$#1_{\pm #2}$}
\newcommand{\RNum}[1]{\uppercase\expandafter{\romannumeral #1\relax}}

\begin{document}
\maketitle

\begin{abstract}
Modeling long-range spatiotemporal dynamics in functional Magnetic Resonance Imaging (fMRI) remains a key challenge due to the high dimensionality of the four-dimensional signals. 
Prior voxel-based models, although demonstrating excellent performance and interpretation capabilities, are constrained by prohibitive memory demands and thus can only capture limited temporal windows. 
To address this, we propose TABLeT (Two-dimensionally Autoencoded Brain Latent Transformer), a novel approach that tokenizes fMRI volumes using a pre-trained 2D natural image autoencoder. Each 3D fMRI volume is compressed into a compact set of continuous tokens, enabling long-sequence modeling with a simple Transformer encoder with limited VRAM. Across large-scale benchmarks including the UK-Biobank (UKB), Human Connectome Project (HCP), and ADHD-200 datasets, TABLeT outperforms existing models in multiple tasks, while demonstrating substantial gains in computational and memory efficiency over the state-of-the-art voxel-based method given the same input. 
Furthermore, we develop a self-supervised masked token modeling approach to pre-train {\ours}, which improves the model's performance for various downstream tasks.
Our findings suggest a promising approach for scalable and interpretable spatiotemporal modeling of brain activity.
Our code is available at \url{https://github.com/beotborry/TABLeT}.
\end{abstract}

\input{contents/introduction}

\input{contents/related_works}

\input{contents/method}
\input{contents/experimental_results}
\input{contents/conclusion}

\section*{Acknowledgment}
This work was supported in part by National Research Foundation of Korea (NRF) grant
[No. RS-2025-02263628, No. RS-202300265406], the Institute of Information \& communications Technology Planning
\& Evaluation (IITP) grants [RS-2021-II212068, RS-2022-II220113, RS-2022-II220959, RS-2021-II211343], the BK21 FOUR Education and
Research Program for Future ICT Pioneers (Seoul
National University), funded by the Korean government (MSIT), the Ministry of Education [RS-2024-00435727], the National Supercomputing Center [KSC-2025-CRE-0340], the U.S. Department of Energy’s ASCR Leadership Computing Challenge [m4750-2024], and Hyundai Motor Chung Mong-Koo Foundation.

{
    \small
    \bibliographystyle{ieeenat_fullname}
    \bibliography{main}
}

\appendix
\input{contents/supplementary}

\end{document}

%% file: contents/introduction.tex
\section{Introduction}

The human brain is a spatiotemporal dynamic system whose activity can be non-invasively measured using functional magnetic resonance imaging (fMRI). 
A large body of work has leveraged fMRI to investigate functional connectivity patterns for tasks such as neurological disorder diagnosis or demographic attribute prediction \citep{BNC, BNT, meanmlp, TFF, swift, brainlm, brainjepa}. 
Existing approaches can be broadly divided into two categories: ROI-based methods and voxel-based methods.  
\begin{figure}
  \centering
  \includegraphics[width=\linewidth]{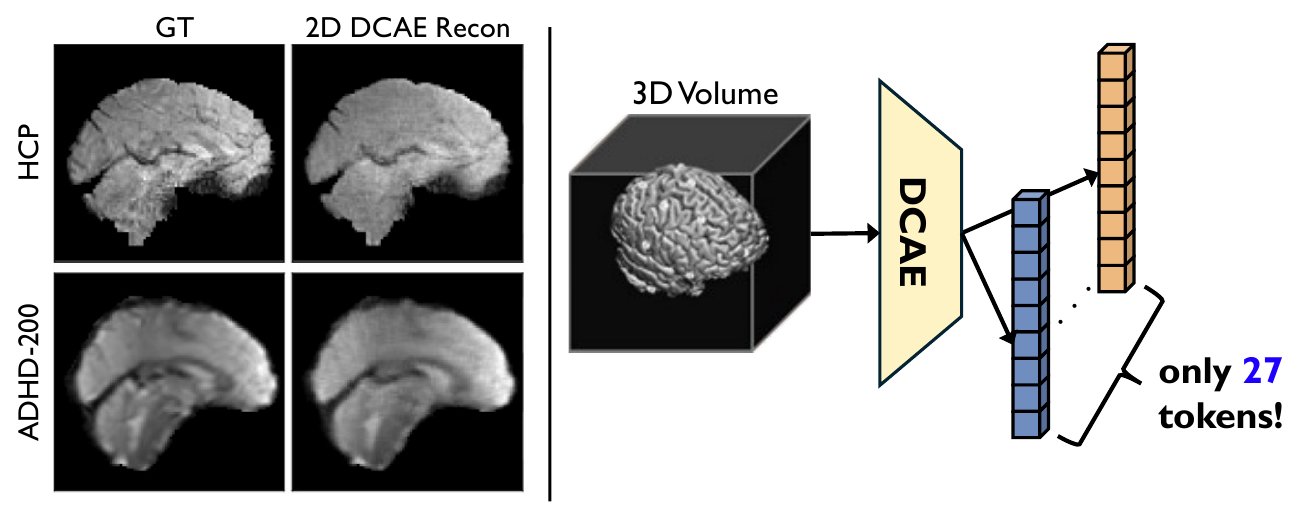}
  \caption{We show that a 2D natural image autoencoder can be transferred well to 4D fMRI data (left). Building on this, we propose to tokenize a single 3D volume of fMRI only into 27 tokens, enabling longer spatiotemporal modeling with a simple Transformer architecture (right).}
  \label{fig:teaser}
  \vspace{-5mm}
\end{figure}
ROI-based methods first define a set of regions of interest (ROIs) based on anatomical segmentation \citep{FC}, extract their corresponding time-series signals, and then compute functional connectivity (FC) matrices as model inputs.
Although this approach is computationally efficient for managing the high dimensionality of fMRI data, it has several limitations: performance strongly depends on the choice of ROIs, fine-grained 3D spatial structures may be lost, and aggressive compression can discard informative signals. 
To overcome these limitations, voxel-based methods such as TFF~\citep{TFF} and SwiFT~\citep{swift} have been proposed. These methods directly process raw 4D fMRI data, thereby preserving spatial and temporal information, while also allowing detailed interpretation as they directly operate on the given image. 
However, due to the massive scale of fMRI volumes, the temporal length that could be simultaneously processed by the model is severely restricted (e.g., TFF and SwiFT use only 20 timesteps at once), potentially missing informative long-range temporal dynamics, and limiting use for tasks that require longer-range interactions, such as the infraslow BOLD–LFP coupling and global arousal waves that unfold over tens of seconds~\citep{pan2013infraslow,raut2021global}.

In this work, we aim to improve voxel-based fMRI modeling by \emph{tokenizing} fMRI volumes into a compact set of continuous tokens, thereby enabling Transformers \citep{attention} to process substantially longer temporal sequences. 
To this end, we observed a strong perceptual information preservation capability of the Deep Compression Autoencoder (DCAE)~\citep{DCAE} and aimed to leverage it, as it effectively tokenizes a \(256 \times 256\) 2D natural image into just 64 continuous tokens (a compression ratio of 32). 
Motivated by this, we ask whether a high-performing \textit{\textbf{2D}} autoencoder trained on \textit{\textbf{natural images}} can serve as an effective tokenizer for \textit{\textbf{4D fMRI}} data. 

Although it may seem counterintuitive, we demonstrate that such an autoencoder can effectively tokenize fMRI volumes while preserving both fine-grained and high-level functional information.
Namely, by rearranging the tokens extracted from each 2D slice of a 3D fMRI volume, we compress an entire volume into \textit{only} 27 continuous tokens, thereby dramatically reducing the input size and enabling substantially longer-sequence modeling with a simple Transformer encoder-based architecture even with limited VRAM. We dub our method {\ours}, \textbf{T}wo-dimensionally \textbf{A}utoencoded \textbf{B}rain \textbf{L}at\textbf{e}nt \textbf{T}ransformer, which achieves consistent yet modest improvements compared to voxel-based baselines on demographic attribute prediction and attention-deficit hyperactivity disorder (ADHD) diagnosis tasks. 

Our model drastically saves memory and computation costs compared to the most efficient voxel-based baseline under the same input data scale, allowing much longer sequences to be fed into the model. This unlocks the opportunity for future use cases that require such long sequences.
Moreover, we demonstrate that self-supervised masked token modeling on fMRI data can effectively pre-train the model, thereby improving downstream task performance.

Our core contributions are summarized as follows:

\begin{compactitem}
\item We propose a novel tokenization scheme that utilizes a 2D autoencoder, which is pre-trained on natural images. We then empirically demonstrate that, despite the apparent domain mismatch, it can serve as a practical general-purpose tokenizer for fMRI signals.
\item Building on this tokenization, we construct a simple Transformer architecture to model substantially longer temporal sequences than prior state-of-the-art voxel-based models, leading to performance improvements across multiple tasks based on resting-state fMRI (rs-fMRI) datasets. This approach would allow application to tasks that require longer-range interactions.
\item Furthermore, we show the feasibility of pre-training on tokenized fMRI data using self-supervised masked token modeling, which can further improve downstream task performance.
\end{compactitem}

%% file: contents/related_works.tex
\section{Related Work}
\begin{figure*}[t!]
  \centering
  \begin{subfigure}[t]{0.49\linewidth}
    \includegraphics[width=\linewidth]{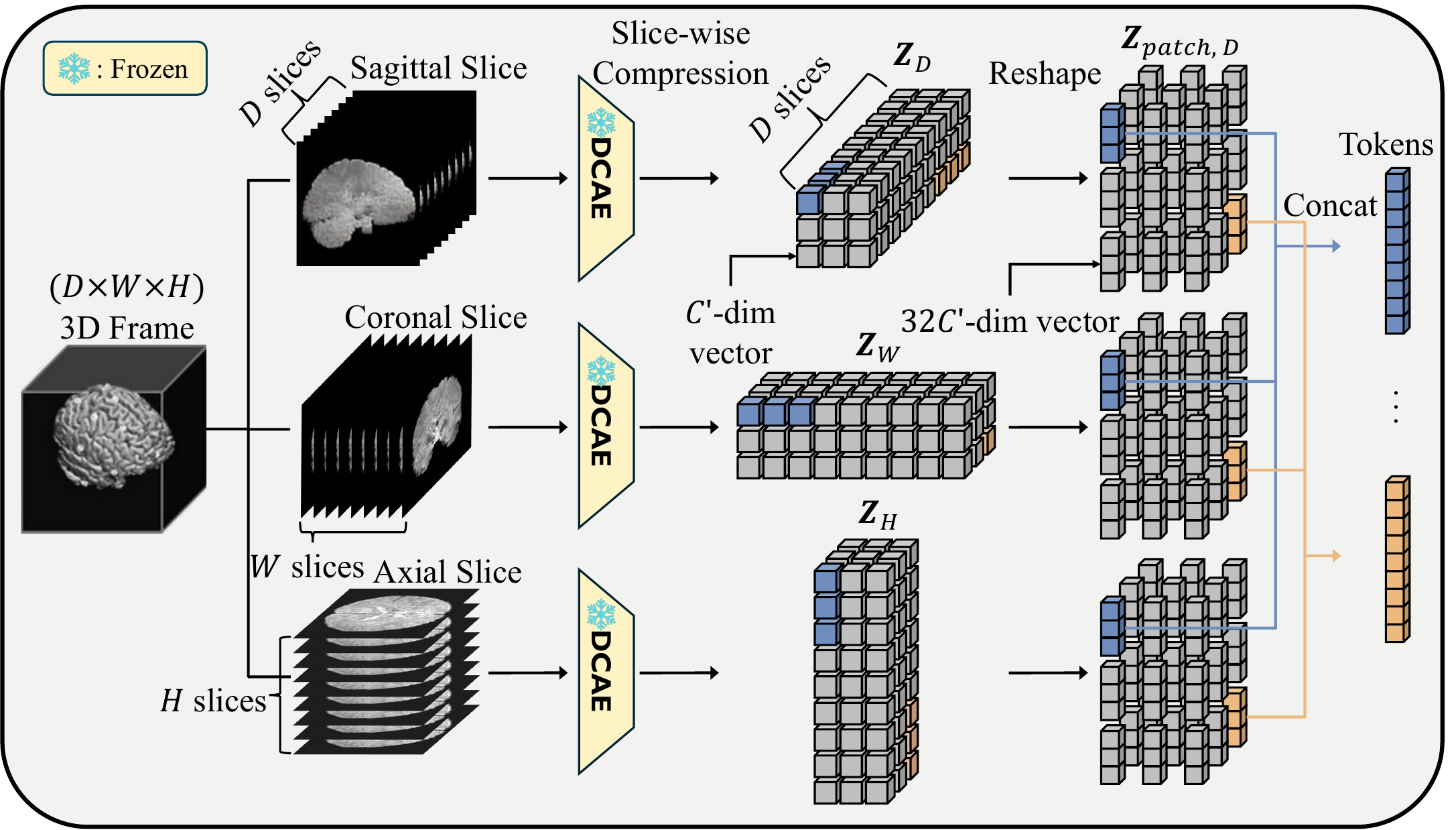}
    \caption{Tokenization process of a 3D volume using a 2D encoder.}
  \end{subfigure}\hfill
  \begin{subfigure}[t]{0.49\linewidth}
    \includegraphics[width=\linewidth]{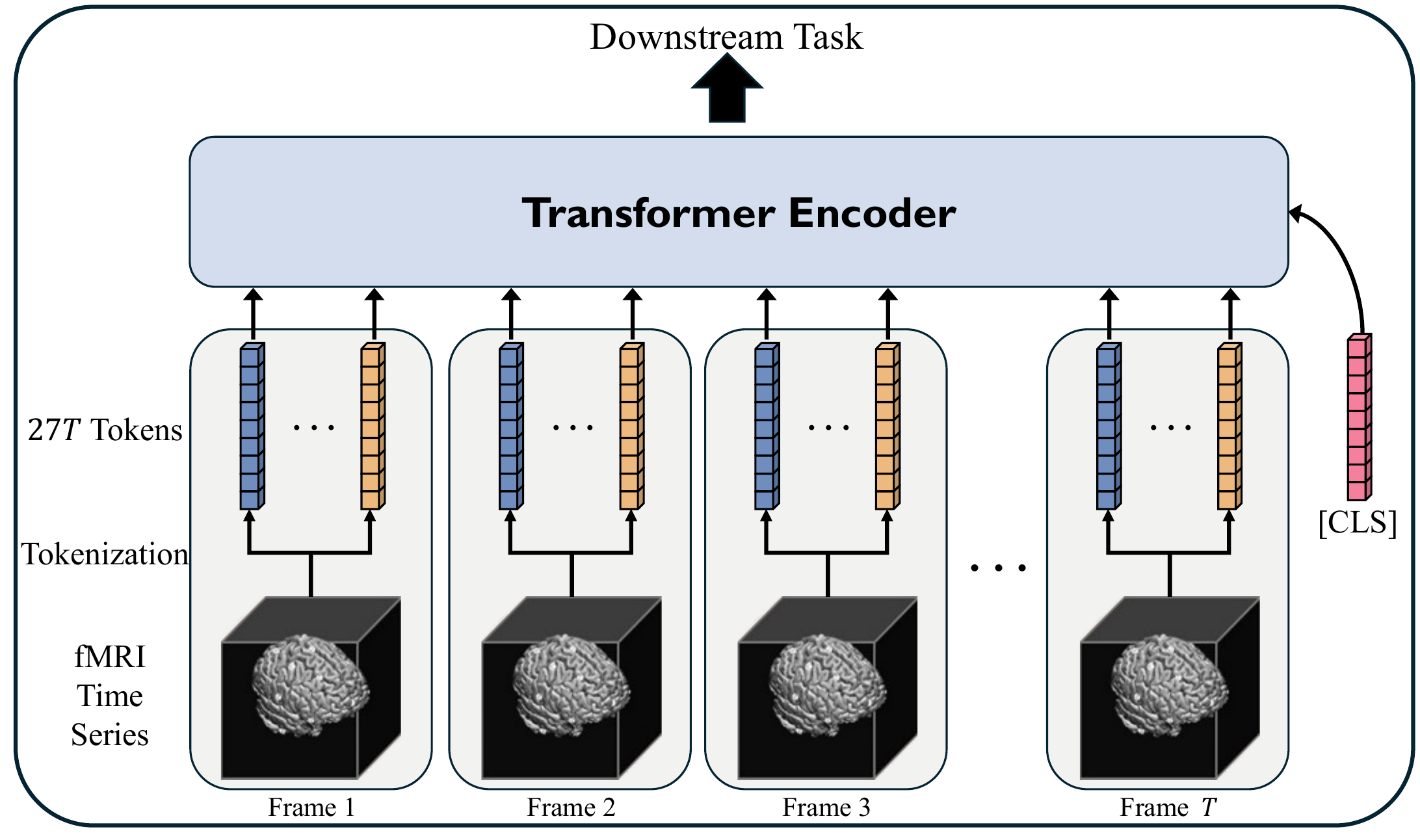}
    \caption{Overview of {\ours}.}
  \end{subfigure}
  \caption{In {\ours}, each frame of the fMRI timeseries is tokenized by a 2D autoencoder, and the tokens are processed by a Transformer.}
  \label{fig:overview}
  \vspace{-3mm}
\end{figure*}
\label{sec:related_works}
\noindent \textbf{ROI-Based Methods.} ROI-based methods parcellate the brain into ROIs and average the BOLD signals within each. The signals are transformed into FC matrices by computing the correlation between the time series of the ROIs.
BrainNetCNN \citep{BNC} treats the FC matrix as a 2D image and uses edge-to-edge, edge-to-node, and node-to-graph convolutional filters to utilize topological locality in ROI-based networks. 
Brain Network Transformer \citep{BNT} adapts the Transformer architecture to process FC matrices as graphs of ROIs. 
meanMLP \citep{meanmlp} is a lightweight MLP-based model that applies the same MLP repeatedly across parcellated fMRI time-series and averages the resulting embeddings across time before a final classification layer. 
Brain-JEPA \citep{brainjepa} is a joint-embedding predictive architecture (JEPA) model pretrained on parcellated fMRI with spatiotemporal masking and gradient-based positioning. Even though computationally efficient, they are inherently limited by the strong pre-processing step that turns brain signals into FC matrices; it is heavily influenced by the choice of ROIs, and during the process, structural information as well as other signals can be discarded.

\noindent \textbf{Voxel-Based Methods.} Voxel-based methods process 4D fMRI volumes, enabling end-to-end learning of spatiotemporal features without ROI aggregation.
TFF \citep{TFF} operates on entire 4D volumes using a two-phase approach: self-supervised pretraining to reconstruct 3D volumes and fine-tuning. It captures fine-grained spatiotemporal dynamics, enabling transfer learning from unlabeled data.
SwiFT \citep{swift} extends the Swin Transformer to 4D fMRI volumes with a 4D window multi-head self-attention mechanism and absolute positional embeddings. 
Voxel-based methods are free from the issues with ROI-based models; however, they are burdened with a higher memory and computation load, as they need to deal with high-dimensional data.

\noindent \textbf{Self-Supervised Pretraining.}
Self-supervised learning (SSL) has emerged as a powerful pre-training framework for vision models, enabling scalable representation learning from unlabeled datasets.
MAE \citep{mae} introduces an asymmetric encoder-decoder architecture, where a high portion of input image patches are masked. Visible patches are encoded by Vision Transformer (ViT), and the decoder reconstructs the masked patches in pixel space. MAE demonstrated superior downstream task transfer, such as classification and segmentation.
SimMIM \citep{simmim} proposes a masked image modeling (MIM) framework using hierarchical transformers and a simple linear prediction head.
VideoMAE \citep{videomae} extends the MAE framework to videos by randomly applying masks on spatio-temporal cubes across spatio-temporal dimensions. The model reconstructs the masked cubes, learning dynamics, and long-range interactions.

\noindent \textbf{Deep Compression Autoencoder.}
Deep Compression Autoencoder (DCAE) \citep{DCAE} introduces an autoencoder framework for accelerating high-resolution diffusion models through extreme spatial compression ratios of up to 128$\times$. DCAE achieves superior reconstruction quality at high compression levels by residual autoencoding. Residual autoencoding utilizes non-parametric shortcuts that enable the model to learn residuals. The encoder downsample blocks adapt a space-to-channel operation, and the decoder upsample blocks use a channel-to-space operation. These non-parametric operations effectively preserve information without learned parameters.

%% file: contents/method.tex
\section{Method}

\subsection{Tokenization of fMRI with 2D Natural Image Autoencoder}
\label{sec:3.1}
To enable voxel-based, long-range fMRI dynamics modeling, our goal was to design a tokenizer that could substantially compress fMRI voxels while minimizing information loss.
We chose to employ the encoder part of an autoencoder, since it can handle tokenization while preserving coarse spatial topology. 
A straightforward strategy would be training an autoencoder directly on fMRI data. However, this approach is both computationally prohibitive and data hungry because training requires large sample sizes that are rarely available in medical imaging. Moreover, the resulting models often generalize poorly, as fMRI characteristics vary across scanners and acquisition protocols.

Therefore, we propose to directly use an autoencoder trained with natural images, without the need for further fine-tuning on medical data.
Among existing options, we adopt DCAE, which achieves high compression rates while maintaining reconstruction fidelity.

To check if it would be feasible to use off-the-shelf autoencoders to tokenize fMRI data, we first measured how much information would be preserved during tokenization. We compared an off-the-shelf 2D natural image DCAE (hereafter 2D DCAE) with a 3D DCAE trained directly on fMRI (hereafter 3D DCAE), as detailed in Sec.~\ref{sec:recon_quality}.
We note that fMRI data are timeseries of 3D images, while the 2D DCAE only operates with 2D images. Therefore, we slice the data into 2D images and feed them into the autoencoder. 

The 2D DCAE showed comparable information preservation capabilities across both fine-grained spatial details and global functional patterns.
Based on this finding, we propose to tokenize each 3D volume independently using the 2D DCAE encoder, in a slice-wise manner, and apply this procedure across the entire fMRI sequence, as described below.

\noindent \textbf{Tokenization of a 3D Volume with Slicing.}  
Each fMRI frame is a 3D volume $\mathbf{X} \in \mathbb{R}^{1 \times D \times H \times W}$. The single channel is first duplicated across three channels to simulate an RGB structure, giving $\mathbf{X} \in \mathbb{R}^{3 \times D \times H \times W}$. One spatial dimension is then chosen as the slicing axis, so the volume becomes a stack of 2D images. For example, if we slice by the depth axis, the volume is treated as $D$ images of shape $\mathbb{R}^{3 \times H \times W}$. Each image slice is compressed independently into a latent representation $\mathbf{Z} \in \mathbb{R}^{C' \times \frac{H}{32} \times \frac{W}{32}}$, where the factor of 32 is the DCAE's spatial compression ratio.  

\noindent \textbf{Aggregation of 3 Axes.}  
This procedure is repeated for all three slicing axes, producing three latent volumes: $\mathbf{Z}_D \in \mathbb{R}^{D \times C' \times \frac{H}{32} \times \frac{W}{32}}, \quad
\mathbf{Z}_H \in \mathbb{R}^{H \times C' \times \frac{D}{32} \times \frac{W}{32}}, \quad
\mathbf{Z}_W \in \mathbb{R}^{W \times C' \times \frac{D}{32} \times \frac{H}{32}}$. As we want to align the three possible latent volumes, the latents are grouped and concatenated along the uncompressed dimension (the slicing axis) by patches of size 32, reshaping them to $\mathbf{Z}_{\text{patch},D}, \mathbf{Z}_{\text{patch},H}, \mathbf{Z}_{\text{patch},W} \in \mathbb{R}^{32C' \times \frac{D}{32} \times \frac{H}{32} \times \frac{W}{32}}$.
This yields $\frac{D}{32} \times \frac{H}{32} \times \frac{W}{32}$ tokens for each slicing variation, where each token corresponds to a position in the downsampled 3D grid and has a hidden dimension $32C'$. The tokens from the three variations are then concatenated with the tokens from other variations that belong to the same spatial position, resulting in $\frac{D}{32} \times \frac{H}{32} \times \frac{W}{32}$ tokens per frame with hidden dimension $96C'$. In our case, $H = W = D = 96$ and $C' = 32$, which means each 3D volume of shape $(1, 96, 96, 96)$ is tokenized into 27 tokens with an embedding dimension of 3072. 
Finally, we note that tokenization is performed only once, and the tokens are cached for later use, making its computational cost negligible compared to the subsequent training process.
We also investigate different aggregation schemes in \cref{sec:variation_tokenization}, which shows the downstream performance is not majorly affected by the specific aggregation scheme.

\subsection{{\ours} Model Architecture}
To capture the spatiotemporal dynamics of tokenized fMRI sequences, we design a simple yet effective Transformer encoder \citep{attention}, naming the pipeline \textbf{TABLeT} (Two-dimensionally Autoencoded Brain Latent Transformer).
The architecture is built on a standard Transformer encoder backbone and integrates several modern components commonly adopted in large language models \citep{qwen2.5, llama3}. In particular, we adopt grouped query attention \citep{GQA} to efficiently handle long sequences, along with the rotary positional encoding \citep{RoPE}.
In addition, we employ \texttt{F.scaled\_dot\_product\_attention} from PyTorch \citep{pytorch}, which offers both speed and memory savings.
Before being fed into the Transformer, fMRI tokens are normalized and projected into a lower-dimensional embedding space via a linear layer. A \texttt{[CLS]} token is prepended to the sequence, followed by an additional normalization step to enhance training stability.
Unless stated otherwise, the model is composed of 12 Transformer layers with 14 attention heads and 2 key–value heads, processing sequences of tokens from 256 frames at once ($T=256$).
We randomly sampled 256 frames from the entire sequence of each subject at every training iteration, while for validation, we used all of the frames by partitioning the sequence and averaging the outputs across partitions, following \citet{swift}.

\subsection{Self-supervised Pre-training with Masked Token Modeling}
\label{sec:masked_pretrain}
Taking inspiration from masked image modeling (MIM) approaches like SimMIM \citep{simmim}, we propose a masked \textit{token} modeling (MTM) approach to pre-train the Transformer encoder of {\ours}. The key difference is that instead of masking image pixels or patches, we choose to directly mask the tokenized representations of the image;
starting from the tokens created by the 2D DCAE, we randomly mask some of the tokens by replacing them with \texttt{[MASK]} tokens. From the partially masked input tokens, we task the Transformer encoder to predict the masked tokens by passing the output tokens through a linear prediction head that reconstructs the input tokens. The model is trained through an $\mathcal{L}_1$ loss exclusively on the masked tokens:

\vspace{-0.1in}
\begin{equation}
    L = \frac{1}{\Omega(\mathbf{Z}_M)}||\mathbf{y}_M-\mathbf{Z}_M||_1,
\end{equation}
in which $\mathbf{Z}, \mathbf{y} \in \mathbb{R}^{96C' \times \frac{D}{32} \times \frac{H}{32} \times \frac{W}{32}}$ are the input tokens and the predicted tokens, respectively. The subscript $M$ denotes the set of masked tokens, and $\Omega$ counts the number of elements (thus the number of masked tokens). We used a masking ratio of 0.5 as the default in our experiments.

Even though it would be possible to use a typical MIM approach by masking the brain volume directly, we chose MTM as it streamlines the pre-training process by removing the need to include the image encoder and decoder.

\noindent \textbf{Masking Strategy.}
We mask the input tokens with a learnable mask token.
Also, instead of masking the tokens in a completely random manner, the same masking pattern from a single frame is repeated across different frames, similar to the tube masking strategy found within VideoMAE \citep{videomae}. This is a measure to prevent the model from ``cheating" by looking at unmasked tokens in the same location from different frames.

%% file: contents/experimental_results.tex
\section{Experimental Results}
\input{tables/main_results}
\input{tables/pretraining}

\label{sec:experimental results}

\subsection{Experimental Setting}
\label{sec:4.1}
\noindent \textbf{Datasets.}
We used resting-state fMRI data from 8,178 middle-aged and older adults from UK-Biobank (UKB) \citep{UKB}, from 1,061 healthy young adults in the Human Connectome Project (HCP) \citep{HCP}, and from 533 children and adolescents, including both individuals diagnosed with ADHD and healthy controls, included in ADHD-200 \citep{ADHD}.

For UKB and HCP, we used the preprocessed data provided by UK-Biobank \citep{UKB_preproc1, UKB_preproc2} and  HCP \citep{HCP}, which go through the preprocessing pipeline including bias field reduction, skull-stripping, cross-modality registration, and spatial normalization to the MNI space \citep{MNI152}.
For ADHD-200, we used the fMRIPrep \citep{fmriprep1, fmriprep2} processed data from \citet{ADHD} and regressed out nuisance variables using cosine bases, six motion parameters, and aCompCor components.
Following \citet{swift}, we set each fMRI volume to the shape of $(96,96,96)$ by cropping out the background and padding appropriately, and we then applied global min-max normalization and rescaled it to the [$-$1, 1] range to match the input range of the DCAE.

We split UKB and HCP using stratified sampling: by age and sex for UKB, and by age, sex, and intelligence score for HCP.
For the ADHD-200 dataset, we performed stratified sampling based on diagnosis labels and image acquisition sites, following \citet{BNT}.
We generated four different random stratified splits, and for all of the splits, the training, validation, and test sets were assigned in a $0.7:0.15:0.15$ ratio. For the ADHD-200 dataset, we experimented with three random training seeds for each split to ensure reliable results, given the relatively small size of the dataset.

\noindent \textbf{Prediction Targets and Evaluation Metrics.}
We considered sex and age for both UKB and HCP, intelligence (\texttt{CogTotalComp-AgeAdj}) for HCP, and diagnosis for ADHD-200. The continuous targets (age, intelligence) are z-normalized using the training set.
Classification tasks were evaluated with accuracy, AUC (Area Under ROC Curve), and F1 score. Regression tasks were evaluated with MAE (Mean Absolute Error), MSE (Mean Squared Error), and $\rho$ (Pearson’s correlation).

\noindent \textbf{Baselines.}
We considered five ROI-based models as our baseline: XGBoost (eXtreme Gradient Boosting) \citep{xgboost} as a standard machine learning baseline, BrainNetCNN \citep{BNC}, Brain Network Transformer (BNT) \citep{BNT}, meanMLP \citep{meanmlp}, and Brain-JEPA \citep{brainjepa}.
For the Brain-JEPA, we trained a model from scratch for a fair comparison.
To preprocess the data, we first construct the FC matrix using a total of 450 ROIs, comprising 400 ROIs from the Schaefer-400 atlas~\citep{schaefer} and 50 additional ROIs from the Tian-Scale \RNum{3} atlas~\citep{tian}.
For the XGBoost model, we used the upper-triangular FC matrix as the input.
We followed the preprocessing pipeline of Brain-JEPA for its experiments.

For the voxel-based baselines, we adopted TFF \citep{TFF} and SwiFT \citep{swift}, the state-of-the-art voxel-based model. We reproduced the original model with 20 input time frames ($T=20$) for both of them. We also extended SwiFT to our hardware limit ($T=50$) to test possible gains from a longer temporal context; alongside the input time frames, the temporal window size was also extended from $4$ to $10$.

\footnotetext{Since the ADHD-200 dataset contains fMRI data with varying repetition time (TR) values and fewer than 160 frames, the default frame number used in Brain-JEPA, we were unable to conduct experiments.}

\subsection{Main Results}
\label{sec:main_results}
Tab.~\ref{tab:main_results} presents experimental results comparing the performance of different models on a training-from-scratch setting, and the standard deviations are detailed in Sec.~\ref{sec:detailed_experimental_results}. 
We also compare the performance between SwiFT and {\ours} while matching $T$ in \cref{sec:matchted_t} to compare the two models in detail.

The results demonstrate that {\ours} outperforms baseline methods, including both ROI-based and voxel-based approaches, across four tasks and three datasets, with only marginal gains on the HCP-Age task and competitive performance 
on the UKB-Sex task.

Interestingly, the results of SwiFT ($T=20,50$) and {\ours} indicate a positive association between the number of input time frames ($T$) and performance, in intelligence prediction and ADHD diagnosis, suggesting that modeling longer temporal variability may be particularly advantageous for these tasks. Sec.~\ref{sec:ablation_studies} expands on this observation with a more detailed study.

Although the overall improvement over SwiFT appears modest, we believe this is likely due to two factors: first, the nature of rs-fMRI, in which signals are recorded while subjects are at rest; and second, the tasks considered in our study, which primarily rely on structural information and may not benefit from longer temporal dependencies.

We expect that applying {\ours} to task fMRI data, which contain more temporally dynamic signals, would result in more substantial performance gains over the baselines as it can cover much longer temporal sequences. 
To support this, we provide preliminary experimental results on HBN-Movie \cite{HBN} in \cref{sec:hbn}, where each subject watches two different movies during fMRI acquisition. The results suggest that {\ours} benefits more from extended temporal context compared to the baselines, demonstrating its potential advantage in settings with highly dynamic stimuli.
In addition, investigating a broader range of tasks to determine which ones can benefit from longer temporal information may provide valuable scientific insights. We believe these are promising directions for future work.

\subsection{Effect of Pre-training on Downstream Tasks}
Tab.~\ref{tab:pretraining_results} shows the effectiveness of the masked token pre-training strategy described in Sec.~\ref{sec:masked_pretrain}.
We first pre-trained {\ours} on a large UKB dataset with a 9:1 training and validation split. We then fine-tuned the model on HCP to simulate a transfer learning setting. 
For fine-tuning, we only used 10 epochs, which is considerably fewer compared to training from scratch.
The results demonstrate that the pre-training of {\ours} indeed contributes to the improvement of downstream task performance, albeit with varying amounts of success depending on the task.
\label{sec:pretraining_results}
\input{tables/3D_DCAE_comparison}

\subsection{Comparison of 2D Natural Image DCAE and 3D fMRI-trained DCAE}

\label{sec:recon_quality}

To compare the quality of the tokens generated by 2D DCAE and 3D DCAE, we evaluate them in two different aspects: reconstruction quality and downstream performance. The reconstruction quality would serve as a proxy of the crucial information preserved in the latent representations, while the downstream performance would represent the practical impact of the two.

\noindent \textbf{Reconstruction quality.}
We first compare the quality of the fMRI reconstructions from the two autoencoders.
The reconstructions were assessed in two different levels: first, we quantified voxel-level reconstruction fidelity using PSNR, which reflects the preservation of fine-grained intensity details, and SSIM, which captures structural similarity and local spatial patterns. Second, we evaluated the preservation of high-level functional information by comparing the difference in FC matrix between the original fMRI data ($\text{FC}_{\text{orig}}$) and its reconstruction ($\text{FC}_{\text{recon}}$): $||\text{FC}_{\text{orig}}-\text{FC}_{\text{recon}}||_F$.

\begin{figure}
  \centering
  \includegraphics[width=\linewidth]{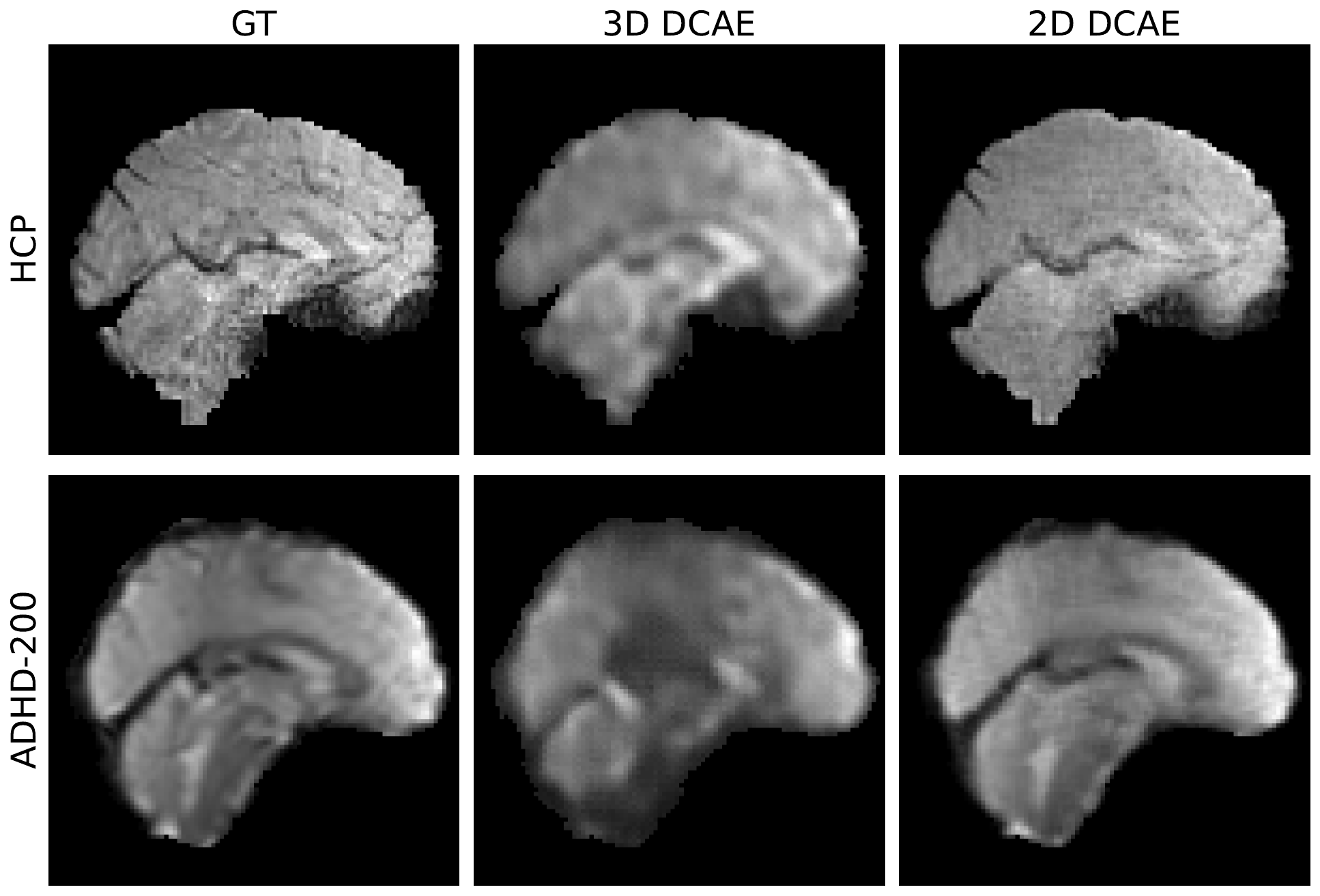}
  \caption{Visualization of reconstructions from 3D and 2D DCAE.}
  \label{fig:recon_vis}
  \vspace{-0.1in}
\end{figure}

\begin{figure}
  \centering
  \includegraphics[width=0.9\linewidth]{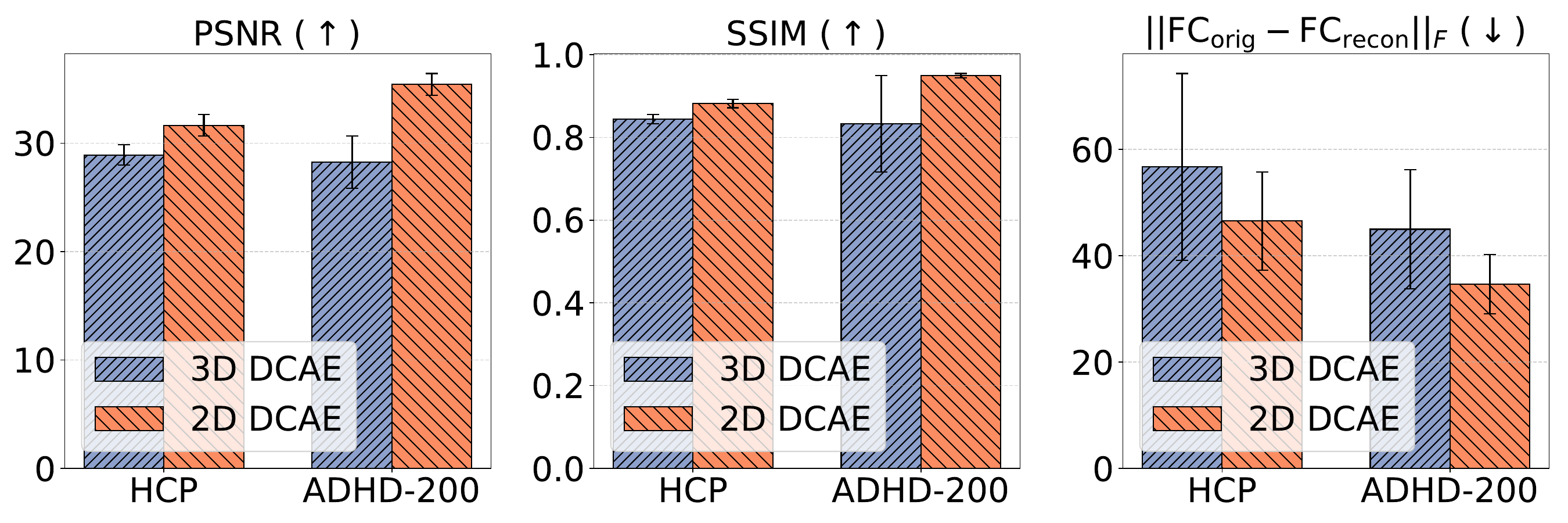}
  \caption{Information preservation of 3D and 2D DCAE.}
  \label{fig:recon_quality}
  \vspace{-5mm}
\end{figure}
\begin{figure*}[ht!]
  \centering
  \begin{subfigure}[t]{0.42\linewidth}
    \includegraphics[width=\linewidth]{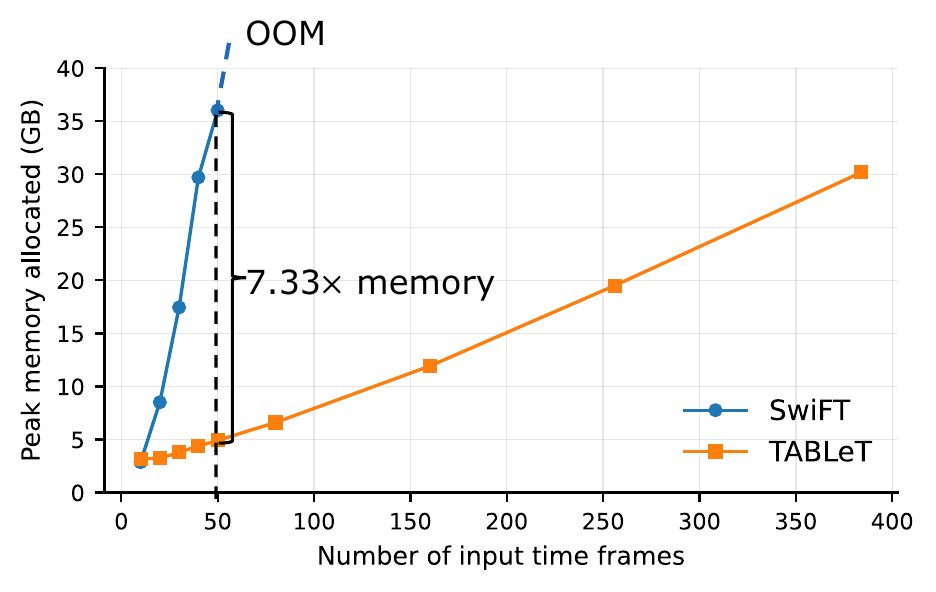}
    \caption{Peak memory allocation}
    \label{fig:left}
  \end{subfigure}
  \begin{subfigure}[t]{0.42\linewidth}
    \includegraphics[width=\linewidth]{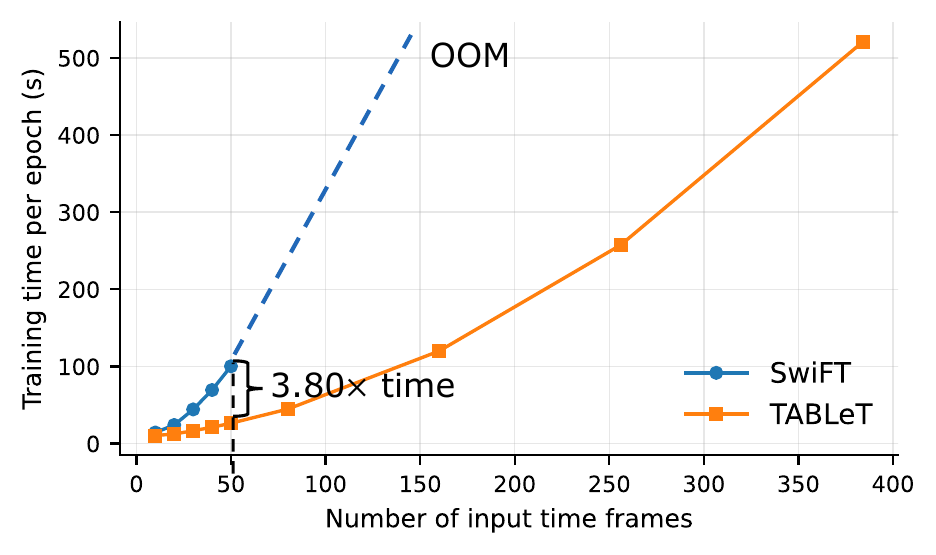}
    \caption{Training time per epoch}
    \label{fig:right}
  \end{subfigure}
  \caption{Comparison of (a) memory and (b) training time, between {\ours} and SwiFT.}
  \label{fig:both}
\end{figure*}
The 3D DCAE was trained with the UKB dataset; a detailed training procedure is provided in Sec.~\ref{sec:training_3D_DCAE}.
To assess generalizability, we deliberately excluded HCP and ADHD-200 from the training set. The reconstructions from the three slicing axes were averaged for 2D DCAE.

The 2D DCAE moderately outperformed the 3D DCAE both qualitatively and quantitatively, as shown in Fig.~\ref{fig:recon_vis} and Fig.~\ref{fig:recon_quality}.
This result suggests that the 2D DCAE, pre-trained on massive, diverse natural image datasets, learns highly robust and general-purpose low-level feature extractors (e.g., for edges, textures), and in turn is able to generalize well into medical images.
Moreover, the 2D DCAE also better preserved functional connectivity patterns, indicating that its latent representations retain high-level functional relationships in the brain.

As a side note, we also attempted to fine-tune the 2D DCAE with fMRI data while freezing different parts of the autoencoder, but discovered that any fine-tuning consistently harmed the reconstruction quality. We presume this is because our fMRI dataset is relatively small and homogeneous compared to the dataset the model is trained on, potentially harming generic filters crucial for the model's generalization capabilities. 

\noindent \textbf{Downstream Performance.}
We also compare {\ours} models trained with latents from the 3D DCAE and the 2D DCAE. The experiment setup is identical to the one detailed in ~\cref{sec:4.1} and ~\cref{sec:main_results}. 
As shown in Tab.~\ref{tab:3D_DCAE_comparison}, both models achieve comparable performance, with the 2D DCAE moderately outperforming the 3D counterpart in most cases.

Overall, these results indicate that a 2D DCAE, despite never being trained on fMRI, generalizes well as a tokenizer for 4D fMRI, both in terms of reconstruction fidelity and downstream performance. Given its strong performance, alongside the practical benefit of avoiding costly dataset collection or additional fine-tuning, we advocate using an existing 2D DCAE as a training-free tokenizer rather than building and training a dedicated 3D counterpart.

\subsection{Memory and Computational Efficiency}

We conduct a quantitative analysis to compare the memory and computational efficiency between {\ours} and SwiFT. To ensure a fair comparison, all tests were performed on a single NVIDIA RTX A6000 GPU, and the batch size of both models was fixed to 4. 
We were only able to run SwiFT up to $T=50$ due to memory limitations. At $T=50$, compared to SwiFT, {\ours} is 7.33 times more memory efficient, and trains 3.8 times faster. With a similar memory budget ($\sim$30GB), $T$ can be extended nearly tenfold between SwiFT ($T=40$) and {\ours} ($T=384$).

\subsection{Additional Ablation Studies}
\label{sec:ablation_studies}
\noindent \textbf{Effect of Aggregation of Three Axes.}
We examined the effect of axis aggregation to better understand its effect: we compared models trained with fMRI tokens derived from a single axis alongside the model with aggregated tokens.
As shown in Tab. \ref{tab:axis_ablation}, the performance of {\ours} varies depending on the chosen axis for single-axis models.
In contrast, our aggregated version consistently achieves strong performance across tasks, eliminating the dependence on any particular slicing axis. We chose to aggregate all three axes instead of using a single axis due to this reliability.
\input{tables/axis_ablation}

\noindent \textbf{Effect of $T$.}
\begin{figure}[ht!]
  \centering
  \begin{subfigure}[t]{0.495\linewidth}
    \includegraphics[width=\linewidth]{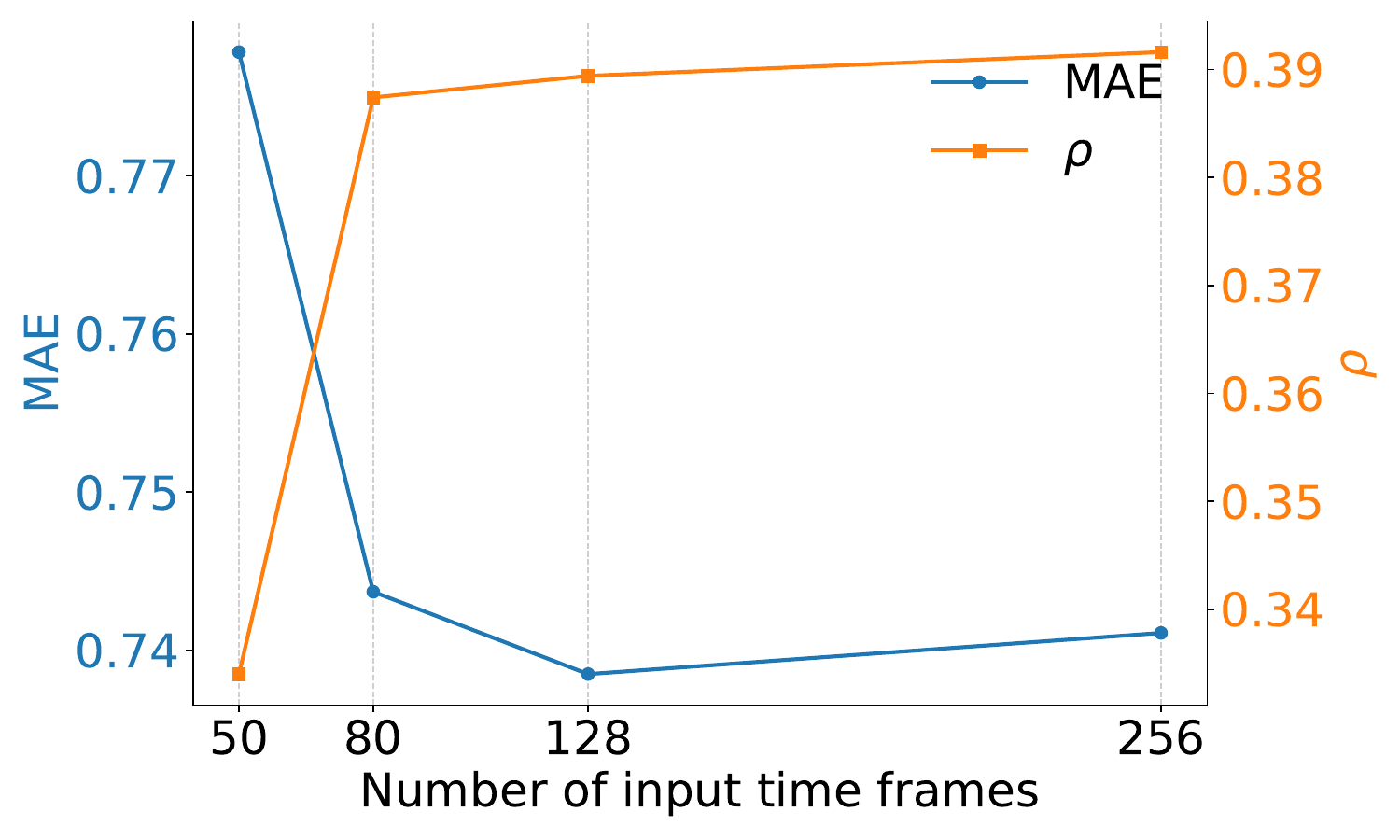}
    \caption{HCP-Intelligence}
    \label{fig:t_left}
  \end{subfigure}
  \begin{subfigure}[t]{0.495\linewidth}
    \includegraphics[width=\linewidth]{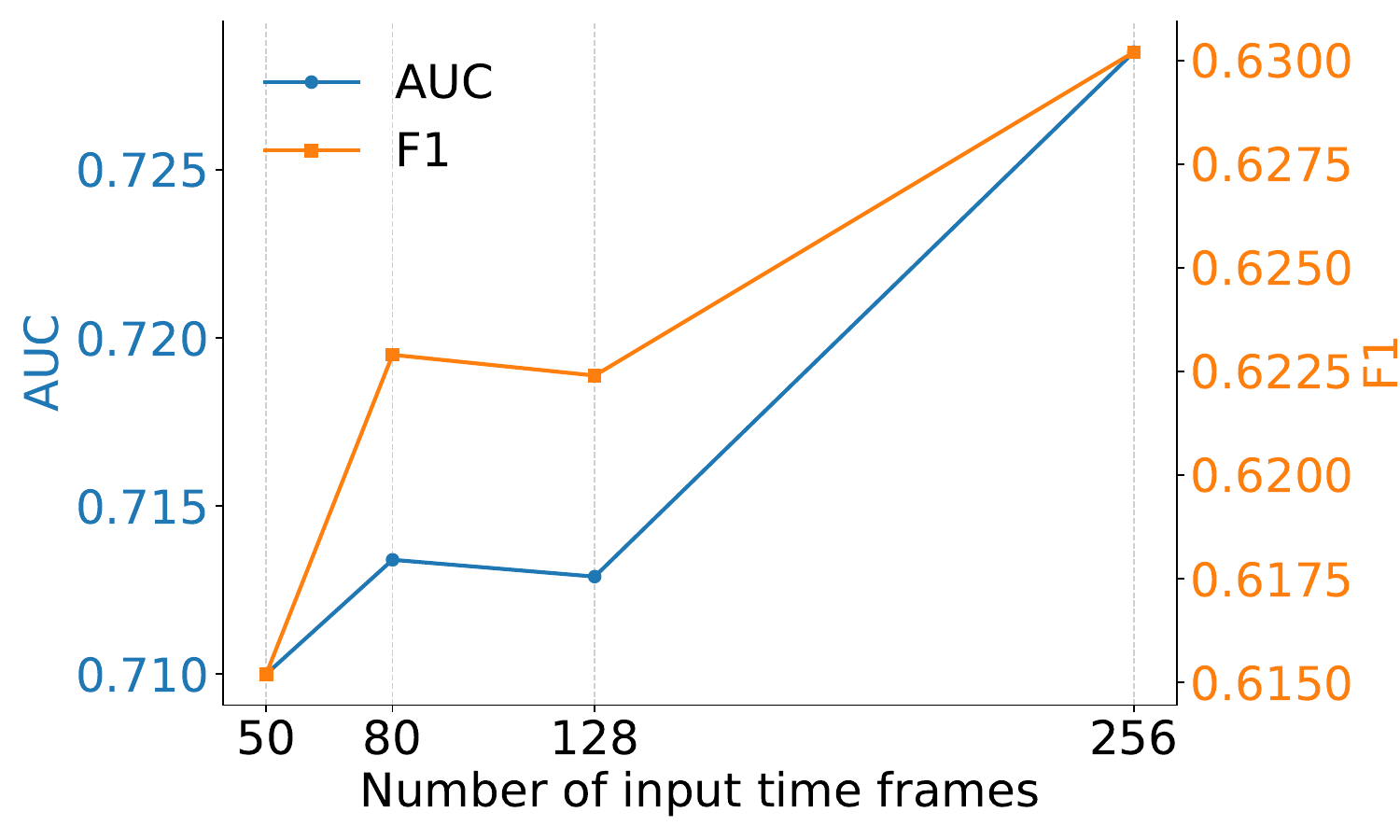}
    \caption{ADHD-200}
    \label{fig:t_right}
  \end{subfigure}
  \caption{Performance of {\ours} on HCP-Intelligence and ADHD-200 with varying $T$.}
  \vspace{-3mm}
  \label{fig:t_ablation}
\end{figure}
As shown in Tab.~\ref{tab:main_results}, modeling longer-range temporal dynamics can improve performance on the HCP-Intelligence and ADHD diagnosis tasks. To explore this further, we varied the $T$ of {\ours} and evaluated the corresponding performance. Interestingly, Fig.~\ref{fig:t_ablation} reveals a clear positive trend between performance and $T$. We believe that investigating the relationship between $T$ and model performance across diverse tasks represents a promising direction for future research.

\subsection{Interpretation Results}

One advantage of voxel-based methods is that the models are interpretable, since the entire process from voxel to prediction is differentiable. To test the interpretability of {\ours},
we used Integrated Gradients (IG) \citep{IG} for visualization of highly contributing areas for sex-classification. We used female test subjects in the HCP-Sex task who are correctly classified with {\ours} with high confidence ($\ge75\%)$, and computed the IG map of the first frame from each subject, then averaged it.
\begin{wrapfigure}{r}{0.2\textwidth}
  \vspace{-0.15in}
  \centering
  \includegraphics[width=\linewidth]{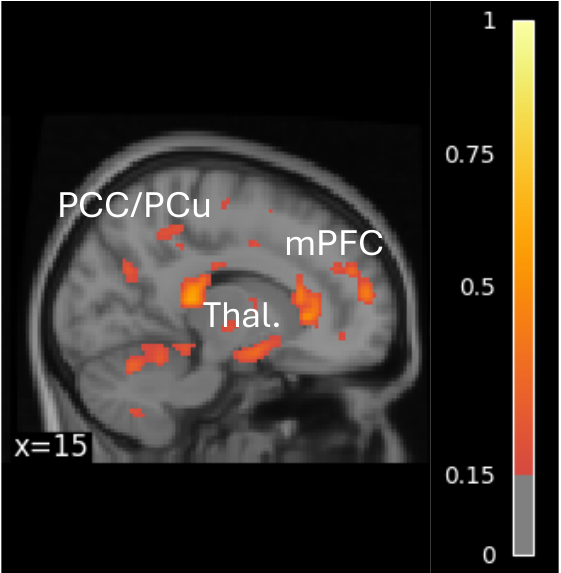}
  \caption{IG map.}
  \label{fig:ig}
  \vspace{-0.3in}
\end{wrapfigure}

Fig.~\ref{fig:ig} shows that {\ours} mainly focuses on the medial prefrontal gyrus (mPFC), posterior cingulate cortex (PCC), precuneus (PCu), and thalamus (Thal.), regions commonly implicated in brain sex difference literature \citep{sexdiff1, sexdiff2, sexdiff3, sexdiff4}.

%% file: tables/main_results.tex
\begin{table*}[!ht]
\caption{Performance comparison to baselines on classification and regression tasks. The best results are \textbf{bolded} and the second best results are \underline{underlined}. Standard deviations are provided in Sec.~\ref{sec:detailed_experimental_results}.}
\centering
\resizebox{.9\linewidth}{!}{
\begin{tabular}{c}
    \begin{minipage}{\linewidth}
    \centering
    \begin{tabular}{l|cccccc|ccc}
    \hline
    \multirow{3}{*}{Method} & \multicolumn{6}{c|}{UKB} & \multicolumn{3}{c}{ADHD-200} \\ \cline{2-10} 
     & \multicolumn{3}{c|}{Sex} & \multicolumn{3}{c|}{Age} & \multicolumn{3}{c}{Diagnosis} \\
     & ACC & AUC & \multicolumn{1}{c|}{F1} & MSE & MAE & \multicolumn{1}{c|}{$\rho$} &  ACC & AUC & F1 \\ \hline
    XGBoost & 84.1 &0.916&\multicolumn{1}{c|}{0.830} & 0.698 & 0.686 & 0.553 & 62.3 & 0.650 & 0.555 \\
    BrainNetCNN & 91.7 & 0.969 & \multicolumn{1}{c|}{0.912} & 0.597 & 0.618 & 0.647 & 59.2 & 0.640 & 0.545 \\
    BNT & 92.4 & 0.980 & \multicolumn{1}{c|}{0.919} & 0.540 & 0.588 & 0.685 & 63.6 & 0.677 & 0.624 \\
    meanMLP &87.7& 0.949& \multicolumn{1}{c|}{0.919}& 0.672& 0.662& 0.586 & 56.8 & 0.617 & 0.532 \\ 
    Brain-JEPA\footnotemark & 86.8 & 0.943 & \multicolumn{1}{c|}{0.862} & 0.688 & 0.669 & 0.574 & -- & -- & -- \\ \hline
    TFF ($T=20$) & \textbf{98.3} & \underline{0.998} & \multicolumn{1}{c|}{\textbf{0.982}} & 0.440 & 0.525 & 0.760 & 63.3 & 0.700 & 0.608 \\
    SwiFT ($T=20$) & 97.4 &\underline{0.998}& \multicolumn{1}{c|}{0.972}& 0.366 & 0.480 & 0.800 & 63.3 & 0.693 & 0.623 \\
    SwiFT ($T=50$) & \underline{98.1} &\textbf{0.999} &\multicolumn{1}{c|}{\underline{0.980}}& \underline{0.364} & \underline{0.477} & \underline{0.802} & \underline{63.9} & \underline{0.701} & \underline{0.627} \\ \hline
    {\ours} ($T=256$) &97.7& \underline{0.998}& \multicolumn{1}{c|}{0.976} & \textbf{0.340} & \textbf{0.466} & \textbf{0.814} & \textbf{65.8} & \textbf{0.729} & \textbf{0.630}  \\ \hline
    \end{tabular}
    \vspace{0.1in}
    \end{minipage}
    \\[1em] 
    \begin{minipage}{\linewidth}
    \centering
    \begin{tabular}{l|ccccccccc}
    \hline
    \multirow{3}{*}{Method} & \multicolumn{9}{c}{HCP}  \\ \cline{2-10} 
     & \multicolumn{3}{c|}{Sex} & \multicolumn{3}{c|}{Age} & \multicolumn{3}{c}{Intelligence} \\
     & ACC & AUC & \multicolumn{1}{c|}{F1} & MSE & MAE & \multicolumn{1}{c|}{$\rho$} & MSE & MAE & $\rho$\\ \hline
    XGBoost & 82.2 & 0.890 & \multicolumn{1}{c|}{0.837} & 0.859 & 0.769 & \multicolumn{1}{c|}{0.296} & 0.908 & 0.779 & 0.292 \\
    BrainNetCNN & 86.3 & 0.937 & \multicolumn{1}{c|}{0.866} & 0.847 & 0.749 & \multicolumn{1}{c|}{0.372} & 0.967 & 0.788 & 0.286 \\
    BNT & 86.3 & 0.935 & \multicolumn{1}{c|}{0.872} & 0.794 & 0.719 & \multicolumn{1}{c|}{0.444} & 0.920 & 0.778 & 0.318 \\
    meanMLP & 84.5 & 0.915 & \multicolumn{1}{c|}{0.855} & 0.846 & 0.751 & \multicolumn{1}{c|}{0.370} & 0.887 &0.767  &0.340   \\ 
    Brain-JEPA & 73.9 & 0.809 & \multicolumn{1}{c|}{0.761} & 0.814 & 0.746 & \multicolumn{1}{c|}{0.369} & 0.959 & 0.799 & 0.171\\ 
    \hline
    TFF ($T=20$) & 88.1 & 0.937 & \multicolumn{1}{c|}{0.892} & 0.888 & 0.779 & \multicolumn{1}{c|}{0.246} & 0.898 & 0.767 & 0.312 \\
    SwiFT ($T=20$) & \underline{93.1} & \underline{0.978} & \multicolumn{1}{c|}{\underline{0.937}} & 0.776 & 0.719 & \multicolumn{1}{c|}{0.450} & 0.940 & 0.782 & 0.297  \\
    SwiFT ($T=50$) & 92.2 & 0.972 & \multicolumn{1}{c|}{0.929} & \textbf{0.764} & \textbf{0.699} & \multicolumn{1}{c|}{\underline{0.460}} & \underline{0.865} & \underline{0.758} & \underline{0.354}  \\ \hline
    {\ours} ($T=256$) & \textbf{93.8} & \textbf{0.987} & \multicolumn{1}{c|}{\textbf{0.943}} & \underline{0.773}  & \underline{0.705} & \multicolumn{1}{c|}{\textbf{0.473}} & \textbf{0.835} & \textbf{0.741} & \textbf{0.392}  \\ \hline
    \end{tabular}
    \end{minipage}
\end{tabular}
}
\label{tab:main_results}
\end{table*}

%% file: tables/pretraining.tex
\begin{table*}[ht]
\caption{Performance comparison between {\ours} trained from scratch (TFS) and fine-tuned after masked pre-training (FT) on HCP. 
}

\centering
\resizebox{.8\linewidth}{!}{
\centerline{

\begin{tabular}{l|ccccccccc}
\hline
\multirow{3}{*}{Model} & \multicolumn{9}{c}{HCP}  \\ \cline{2-10} 
 & \multicolumn{3}{c|}{Sex} & \multicolumn{3}{c|}{Age} & \multicolumn{3}{c}{Intelligence}  \\
 & ACC & AUC & \multicolumn{1}{c|}{F1} & MSE & MAE & \multicolumn{1}{c|}{$\rho$} & MSE & MAE & $\rho$ \\ \hline

{\ours} TFS & \meanstd{93.8}{0.9} & \meanstd{\textbf{0.987}}{0.003} & \multicolumn{1}{c|}{\meanstd{0.943}{0.008}} & \meanstd{0.773}{0.077} & \meanstd{\underline{0.705}}{0.038} & \multicolumn{1}{c|}{\meanstd{\underline{0.473}}{0.053}} & \meanstd{\underline{0.835}}{0.053} & \meanstd{\underline{0.741}}{0.028} & \meanstd{\underline{0.392}}{0.062} \\

{\ours} FT & \meanstd{\textbf{95.3}}{1.3} & \meanstd{\underline{0.986}}{0.005} & \multicolumn{1}{c|}{\meanstd{\textbf{0.958}}{0.011}} & \meanstd{\textbf{0.650}}{0.045} & \meanstd{\textbf{0.655}}{0.024} & \multicolumn{1}{c|}{\meanstd{\textbf{0.552}}{0.032}} & \meanstd{\textbf{0.796}}{0.051} & \meanstd{\textbf{0.732}}{0.028} & \meanstd{\textbf{0.435}}{0.046} \\ \hline

\end{tabular}
}
}

\label{tab:pretraining_results}
\vspace{-3mm}
\end{table*}

%% file: tables/3D_DCAE_comparison.tex
\begin{table*}[ht]
\caption{Performance comparison between {\ours} with latents from 3D DCAE and 2D DCAE on HCP and ADHD-200. Standard deviations are provided in Sec.~\ref{sec:detailed_experimental_results}.
}
\centering
\resizebox{.8\linewidth}{!}{
\centerline{

\begin{tabular}{l|ccccccccc|ccc}
\hline
\multirow{3}{*}{Tokenizer} & \multicolumn{9}{c|}{HCP} & \multicolumn{3}{c}{ADHD-200} \\ \cline{2-13} 
 & \multicolumn{3}{c|}{Sex} & \multicolumn{3}{c|}{Age} & \multicolumn{3}{c|}{Intelligence} & \multicolumn{3}{c}{Diagnosis} \\
 & ACC & AUC & \multicolumn{1}{c|}{F1} & MSE & MAE & \multicolumn{1}{c|}{$\rho$} & MSE & MAE & $\rho$ & ACC & AUC & F1 \\ \hline

3D DCAE & 92.2 & 0.973 & \multicolumn{1}{c|}{0.929} & \textbf{0.767} & \textbf{0.693} & \multicolumn{1}{c|}{\textbf{0.475}} & 0.869 & 0.755 & \multicolumn{1}{c|}{0.387} & \textbf{65.8} & 0.711 & \textbf{0.644}  \\

2D DCAE & \textbf{93.8} & \textbf{0.987} & \multicolumn{1}{c|}{\textbf{0.943}} & 0.773 & 0.705 & \multicolumn{1}{c|}{0.473} & \textbf{0.835} & \textbf{0.741} & \textbf{0.392} &\textbf{65.8} & \textbf{0.729} & 0.630 \\ \hline

\end{tabular}
}
}

\label{tab:3D_DCAE_comparison}
\vspace{0mm}
\end{table*}

%% file: tables/axis_ablation.tex
\begin{table*}[ht]
\centering
\caption{Effect of the choice of slicing axis and aggregation of the three axes on classification and regression tasks. The best results are \textbf{bolded} and the second best results are \underline{underlined}.}
\resizebox{.85\linewidth}{!}{%
\begin{minipage}{\linewidth}
    \centering
    \begin{tabular}{l|cccccc}
    \hline
    \multirow{3}{*}{Axis} & \multicolumn{6}{c}{UKB}  \\ \cline{2-7} 
     & \multicolumn{3}{c|}{Sex} & \multicolumn{3}{c}{Age}  \\
     & ACC & AUC & \multicolumn{1}{c|}{F1} & MSE & MAE & \multicolumn{1}{c}{$\rho$}\\ \hline

    Sagittal & \meanstd{\underline{97.3}}{0.9} & \meanstd{0.996}{0.001} & \multicolumn{1}{c|}{\meanstd{\underline{0.971}}{0.009}} & \meanstd{\underline{0.369}}{0.015} & \meanstd{\underline{0.486}}{0.010} & \multicolumn{1}{c}{\meanstd{\underline{0.796}}{0.008}} \\
    Coronal& \meanstd{97.1}{0.4} & \meanstd{0.996}{0.002} & \multicolumn{1}{c|}{\meanstd{0.969}{0.004}} & \meanstd{0.435}{0.012} & \meanstd{0.525}{0.008} & \multicolumn{1}{c}{\meanstd{0.756}{0.009}}   \\
    Axial & \meanstd{\underline{97.3}}{0.4} & \meanstd{\underline{0.997}}{0.000} & \multicolumn{1}{c|}{\meanstd{\underline{0.971}}{0.004}} & \meanstd{0.410}{0.020} & \meanstd{0.509}{0.014} & \multicolumn{1}{c}{\meanstd{0.771}{0.013}}  \\ \hline

    {All}& \meanstd{\textbf{97.7}}{0.2} & \meanstd{\textbf{0.998}}{0.000} & \multicolumn{1}{c|}{\meanstd{\textbf{0.976}}{0.002}} &  \meanstd{\textbf{0.340}}{0.011} & \meanstd{\textbf{0.466}}{0.010} & \multicolumn{1}{c}{\meanstd{\textbf{0.814}}{0.009}} \\ \hline
    \end{tabular}
    
    \vspace{1em} 

    \centering
    \begin{tabular}{l|cccccc}
    \hline
    \multirow{3}{*}{Axis} & \multicolumn{6}{c}{HCP}  \\ \cline{2-7} 
     & \multicolumn{3}{c|}{Sex} & \multicolumn{3}{c}{Age}  \\
     & ACC & AUC & \multicolumn{1}{c|}{F1} & MSE & MAE & \multicolumn{1}{c}{$\rho$}\\ \hline

    Sagittal & \meanstd{91.3}{3.6} & \meanstd{0.972}{0.017} & \multicolumn{1}{c|}{\meanstd{0.920}{0.033}} & \meanstd{0.783}{0.111} & \meanstd{0.721}{0.041} & \multicolumn{1}{c}{\meanstd{0.458}{0.076}} \\
    Coronal& \meanstd{\underline{93.6}}{1.7} & \meanstd{\underline{0.981}}{0.007} & \multicolumn{1}{c|}{\meanstd{\underline{0.941}}{0.015}} & \meanstd{0.855}{0.053} & \meanstd{0.745}{0.023} & \multicolumn{1}{c}{\meanstd{0.376}{0.048}}   \\
    Axial & \meanstd{92.3}{3.0} & \meanstd{0.979}{0.008} & \multicolumn{1}{c|}{\meanstd{0.930}{0.028}} & \textbf{\meanstd{\textbf{0.748}}{0.056}} & \meanstd{\underline{0.711}}{0.015} & \multicolumn{1}{c}{\meanstd{\underline{0.470}}{0.040}}  \\ \hline

    {All}& \meanstd{\textbf{93.8}}{0.9} & \meanstd{\textbf{0.987}}{0.003} & \multicolumn{1}{c|}{\meanstd{\textbf{0.943}}{0.008}} &  \meanstd{\underline{0.773}}{0.077} & \meanstd{\textbf{0.705}}{0.038} & \multicolumn{1}{c}{\meanstd{\textbf{0.473}}{0.053}} \\ \hline
    \end{tabular}

    \vspace{1em} 

    \centering
    \begin{tabular}{l|cccccc}
    \hline
    \multirow{3}{*}{Axis} & \multicolumn{3}{c}{HCP} & \multicolumn{3}{c}{ADHD-200}  \\ \cline{2-7} 
     & \multicolumn{3}{c|}{Intelligence} & \multicolumn{3}{c}{Diagnosis}  \\
     & MSE & MAE & \multicolumn{1}{c|}{$\rho$} & ACC & AUC & \multicolumn{1}{c}{F1}\\ \hline

    Sagittal & \meanstd{\underline{0.842}}{0.058} & \meanstd{\underline{0.744}}{0.028} & \multicolumn{1}{c|}{\meanstd{\textbf{0.401}}{0.060}} & \meanstd{\textbf{65.8}}{2.3} & \meanstd{\underline{0.715}}{0.026} & \multicolumn{1}{c}{\meanstd{\textbf{0.633}}{0.032}} \\
    Coronal& \meanstd{0.850}{0.057} & \meanstd{0.749}{0.029} & \multicolumn{1}{c|}{\meanstd{0.381}{0.065}} & \meanstd{63.5}{3.1} & \meanstd{0.707}{0.036} & \multicolumn{1}{c}{\meanstd{0.621}{0.040}}   \\
    Axial & \meanstd{0.896}{0.070} & \meanstd{0.773}{0.033} & \multicolumn{1}{c|}{\meanstd{0.309}{0.072}} & \textbf{\meanstd{\underline{64.3}}{2.5}} & \meanstd{0.713}{0.022} & \multicolumn{1}{c}{\meanstd{0.622}{0.034}}  \\ \hline

    {All}& \meanstd{\textbf{0.835}}{0.053} & \meanstd{\textbf{0.741}}{0.028} & \multicolumn{1}{c|}{\meanstd{\underline{0.392}}{0.062}} &  \meanstd{\textbf{65.8}}{3.5} & \meanstd{\textbf{0.728}}{0.020} & \meanstd{\underline{0.630}}{0.038} \\ \hline
    \end{tabular}
\end{minipage}
}
\label{tab:axis_ablation}
\end{table*}

%% file: contents/conclusion.tex
\section{Conclusion \& Limitations}
We introduced TABLeT, a simple and scalable framework that uses a natural-image 2D autoencoder as a training-free tokenizer for fMRI volumes, enabling long-range temporal modeling with a lightweight Transformer. Across three datasets, TABLeT achieved competitive or better performance while offering substantial reductions in memory and computation compared to voxel-based baselines. In addition, we show that masked token pre-training further enhances downstream performance.
Our model is based on a striking result: a 2D autoencoder trained on natural images preserves fMRI structure and functional relationships better than a 3D fMRI-trained autoencoder, enabling efficient tokenization without any domain-specific training. Our model opens the opportunity to handle much longer temporal sequences for future researchers.

Despite these advantages, our study has several limitations.
First, {\ours} tokenizes each frame of the fMRI time series independently. This process may disrupt subtle temporal dynamics. Tokenization strategies that directly incorporate temporal dependencies, especially in tasks where fine-grained dynamics are critical, could be explored in the future.
Second, {\ours} processes all tokens jointly, without explicit modeling of their spatial or temporal structure. Architectures designed to leverage spatial and temporal alignment between tokens may further enhance the ability to capture the spatiotemporal dynamics inherent in fMRI data.
Moreover, as discussed in Sec.~\ref{sec:main_results}, the benefits of longer temporal modeling vary across tasks, and a systematic understanding of which neuroscientific phenomena gain the most remains open.

Nevertheless, we believe our study suggests a promising approach, bridging natural image processing and medical imaging, and enabling scalable, efficient spatiotemporal modeling of brain activity.

%% file: contents/supplementary.tex
\onecolumn
\clearpage
\renewcommand{\thetable}{A\arabic{table}}
\renewcommand\thesection{\Alph{section}}
\setcounter{section}{0}
{
\centering
\Large
\textbf{Can Natural Image Autoencoders Compactly Tokenize fMRI Volumes \\for Long-Range Dynamics Modeling?}\\
\vspace{0.5em}Appendix\\
\vspace{1.0em}
}

\section{Implementation Details}
\label{sec:experimental_details}
For the 2D DCAE, we brought the unmodified \texttt{dc-ae-f32c32-in-1.0} checkpoint provided by \citet{DCAE} for all 2D natural image DCAE experiments.

All experiments were conducted on the NVIDIA A100-40GB and RTX A6000 GPUs. We used \texttt{fp16} mixed precision for the training of all models except for TFF due to NaN error during training.

We used \texttt{BCEWithLogitsLoss} for the classification task, and used \texttt{pos-weight} option for the ADHD task to account for class imbalance.
We used \texttt{L1Loss} for the regression tasks.

For the voxel-based models, TFF, SwiFT, and {\ours}, training was performed by randomly sampling consecutive 3D volumes. For evaluation, following \citet{swift}, we computed the final prediction by averaging the model outputs over all possible windows starting from the first frame.

\paragraph{Shared Settings}
We used the following strategy for all of the experiments, unless explicitly stated.
\begin{compactitem}
\item \texttt{Optimizer}: AdamW using a cosine decay learning rate scheduler, with weight decay of $10^{-2}$.

\item \texttt{Hyperparameter Search}: For the UKB-Sex and HCP-Sex tasks, we searched the hyperparameter based on the validation AUROC for each model.
For the UKB-Age, HCP-Age, and HCP-Intelligence tasks, we searched based on the validation MAE.
For ADHD, we searched based on the validation loss to consider the \texttt{pos-weight} for the class imbalance.

\item \texttt{Early Stopping}: We chose the early-stopped model for the BrainNetCNN, BNT, meanMLP, Brain-JEPA, and TFF by default. As we observed that SwiFT and {\ours} are more stable during training, we report results from the final epoch for all tasks.

\end{compactitem}

\paragraph{XGBoost}
We grid searched for hyperparameter tuning of XGBoost for the following.
\begin{compactitem}
    \item \texttt{Maximum depth}: Chosen between 3 and 5
    \item \texttt{Minimal child weight}: Chosen between 1 and 7
    \item \texttt{Gamma}: Chosen between 0.0 and 0.4
    \item \texttt{Learning rate}: Chosen between 0.05 and 0.3
    \item \texttt{Colsample by tree}: Chosen between 0.6 and 0.9
\end{compactitem}

\paragraph{BrainNetCNN}
We trained BrainNetCNN with the following setup: 
\begin{compactitem}
    \item \texttt{Learning rate}: Chosen between $1\times10^{-6}$ and $2\times10^{-4}$ 
    \item \texttt{Batch size}: 64
    \item \texttt{Epochs}: 100 epochs of training
\end{compactitem}

\paragraph{Brain Network Transformer}
We trained Brain Network Transformer with the following setup: 
\begin{compactitem}
    \item \texttt{Learning rate}: Chosen between $1\times10^{-6}$ and $2\times10^{-4}$ 
    \item \texttt{Batch size}: 64
    \item \texttt{Epochs}: 100 epochs of training
\end{compactitem}

\paragraph{meanMLP}
We trained meanMLP with the following setup: 
\begin{compactitem}
    \item \texttt{Learning rate}: Chosen between $1\times10^{-4}$ and $1\times10^{-2}$ 
    \item \texttt{Batch size}: 32
    \item \texttt{Epochs}: 100 epochs of training
\end{compactitem}

\paragraph{Brain-JEPA}
We trained Brain-JEPA from scratch for fair comparison with the following setup.
\begin{compactitem}
    \item \texttt{Learning rate}: Chose between $1\times10^{-5}$ and $7 \times 10^{-4}$.
    \item \texttt{Batch size}: 16
    \item \texttt{Epochs}: 50 epochs of training
\end{compactitem}

\paragraph{TFF}
We trained TFF with the following setup:
\begin{compactitem}
    \item Phase 1
    \begin{compactitem}
        \item \texttt{Learning rate}: $3\times10^{-3}$ for UKB, ADHD, and $7\times10^{-4}$ for HCP
        \item \texttt{Batch size}: 4
        \item \texttt{Epochs}: 100 epochs of training
    \end{compactitem}

    \item Phase 2
    \begin{compactitem}
        \item \texttt{Learning rate}: $1\times10^{-5}$ for UKB, ADHD, and chosen between $1\times10^{-5}$ and $1\times10^{-6}$
        \item \texttt{Batch size}: 2
        \item \texttt{Epochs}: 50 epochs of training
    \end{compactitem}

    \item Fine-tuning
    \begin{compactitem}
        \item \texttt{Learning rate}: Chosen between $1\times10^{-5}$ and $1\times10^{-6}$ for UKB and ADHD, chosen between $3\times10^{-7}$ and $1\times10^{-6}$ for HCP, 
        \item \texttt{Batch size}: 4
        \item \texttt{Epochs}: 10 epochs of training for UKB-Sex, 20 epochs of training for HCP, UKB-Age, and 30 epochs of training for ADHD.
    \end{compactitem}
\end{compactitem}

\paragraph{SwiFT}
We trained SwiFT with the following setup: 
\begin{compactitem}
    \item \texttt{Learning rate}: Chosen between $1\times10^{-6}$ and $5\times10^{-5}$ 
    \item \texttt{Batch size}: 4
    \item \texttt{Epochs}: 25 epochs of training for UKB, HCP, 30 epochs for ADHD.
\end{compactitem}

\paragraph{{\ours}}
We trained {\ours} with the following setup: 
\begin{compactitem}
    \item \texttt{Learning rate}: Chosen between $3\times10^{-7}$ and $5\times10^{-5}$
    \item \texttt{Batch size}: 4
    \item \texttt{Epochs}: 50 epochs of training for HCP-Sex, HCP-Intelligence, ADHD, 30 epochs for age regression, 15 epochs for UKB-Sex.
\end{compactitem}

\section{Training Details of 3D fMRI-trained DCAE}
\label{sec:training_3D_DCAE}
\begin{figure}[ht]
    \centering
    \includegraphics[width=0.7\linewidth]{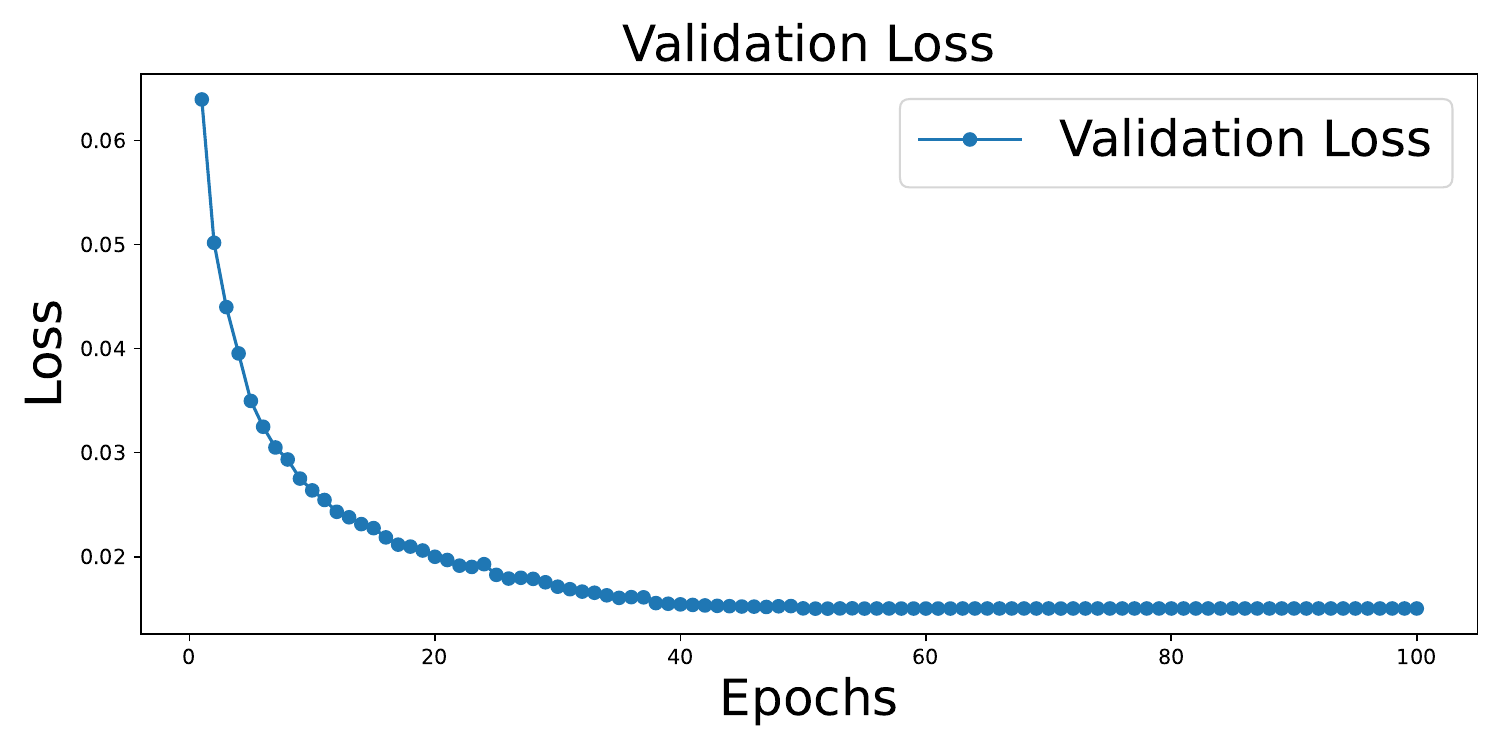}
    \caption{Validation Loss Curve for Training of 3D DCAE.}
    \label{fig:3d_dcae_val_loss}
\end{figure}

We developed 3D DCAE by adapting the architecture of 2D DCAE \citep{DCAE} to handle 3D volume inputs. To achieve this, we replaced 2D convolutional layers with 3D convolutional layers and adjusted components such as RMS normalization, batch normalization, \texttt{PixelUnshuffle}, and \texttt{PixelShuffle} to process 3D data effectively. The model was configured with 1 input channel, 1024 latent channels, encoder-decoder width of \texttt{[16, 64, 256, 256, 1024, 1024]}, and encoder-decoder depth of \texttt{[0, 2, 2, 5, 5, 5]}, to make the same compression ratio as the 2D DCAE.

For training, we utilized a dataset of 8,178 subjects from the UK-Biobank, splitting it into training and validation sets with a 9:1 ratio and stratification based on sex and age. The model was trained for 100 epochs with an initial learning rate of $4\times10^{-5}$, which was gradually reduced using \texttt{ReduceLROnPlateau} scheduler. During each epoch, we randomly selected a single fMRI frame from the full set of frames for each subject to train the model. The training process used $\mathcal{L}_2$ reconstruction loss and the AdamW optimizer with a weight decay of $1\times10^{-4}$.

As the training curve in Fig.~\ref{fig:3d_dcae_val_loss} shows, we made every effort to train the 3D DCAE model to achieve the best performance and ensure full convergence, for fair comparison.

\section{Additional Experimental Results}
\subsection{Experiments with Matched $T$}
\label{sec:matchted_t}
We report the performance of {\ours} with $T=20,50$ in \cref{tab:rebuttal1} on HCP and ADHD-200.
The results demonstrate that {\ours} shows comparable performance to SwiFT. Therefore, it proves that {\ours} is able to maintain competitive performance even with a reduced number $T$, highlighting that the performance gain in \cref{tab:main_results} is not solely from the increased $T$, but rather from the effectiveness of the proposed method itself.

\begin{table}[h!]
\caption{Results of experiments with matched $T$ on HCP and ADHD-200.
}
\centering
\resizebox{\columnwidth}{!}{
\centerline{
\begin{tabular}{l|cccccccccccc}
\hline
\multirow{3}{*}{Model} & \multicolumn{9}{c}{HCP}  & \multicolumn{3}{c}{ADHD-200}\\ \cline{2-13} 
 & \multicolumn{3}{c|}{Sex} & \multicolumn{3}{c|}{Age} & \multicolumn{3}{c|}{Intelligence} & \multicolumn{3}{c}{Diagnosis}  \\
 & ACC & AUC & \multicolumn{1}{c|}{F1} & MSE & MAE & \multicolumn{1}{c|}{$\rho$} & MSE & MAE & \multicolumn{1}{c|}{$\rho$} & ACC & AUC & \multicolumn{1}{c}{F1}\\ \hline
SwiFT $(T=20)$ & \textbf{93.1} & 0.978 & \multicolumn{1}{c|}{\textbf{0.937}} & \textbf{0.776} & 0.719 & \multicolumn{1}{c|}{0.450} & 0.940 & 0.782 & \multicolumn{1}{c|}{0.297} & 63.3 & 0.693 & \textbf{0.623} \\
{\ours \ $(T=20)$} & 91.6 & \textbf{0.980} & \multicolumn{1}{c|}{0.923} & 0.784 & \textbf{0.710} & \multicolumn{1}{c|}{\textbf{0.460}} & \textbf{0.866} & \textbf{0.763} & \multicolumn{1}{c|}{\textbf{0.340}} & \textbf{64.4} & \textbf{0.717} & 0.621 \\
\hline

SwiFT $(T=50)$ & 92.2 & 0.972 & \multicolumn{1}{c|}{0.929} & 0.764 & \textbf{0.699} & \multicolumn{1}{c|}{0.460} & \textbf{0.865} & \textbf{0.758} & \multicolumn{1}{c|}{\textbf{0.354}} & 63.9 & 0.701 & \textbf{0.627} \\
{\ours \ $(T=50)$} & \textbf{93.4} & \textbf{0.986} & \multicolumn{1}{c|}{\textbf{0.940}} & \textbf{0.763} & 0.704 & \multicolumn{1}{c|}{\textbf{0.473}} & 0.916 & 0.778 & \multicolumn{1}{c|}{0.334} & \textbf{64.3} & \textbf{0.710} & 0.615 \\

\end{tabular}
}
}
\label{tab:rebuttal1}
\end{table}

\subsection{HBN-Movie Experiments}
\label{sec:hbn}
To test on a more temporally dynamic task, we experimented on Healthy Brain Network (HBN)-Movie \cite{HBN}, which includes 680 subjects and 1,360 fMRI scans where each subject watched two movies. We trained the models to predict which movie was being viewed. As shown in \cref{tab:rebuttal2}, {\ours} achieves performance comparable to SwiFT with the same temporal length $T$, and outperforms with longer $T$, which demonstrates that {\ours} can handle the temporal dynamics of fMRI data.
\begin{table}[h!]
\caption{Results on HBN-Movie.}
\centering
\resizebox{0.6\linewidth}{!}{
\begin{tabular}{l|ccc}
\hline
\multirow{2}{*}{Model} & \multicolumn{3}{c}{HBN-Movie} \\ \cline{2-4} 
 & ACC & AUC & \multicolumn{1}{c}{F1} \\\hline

SwiFT $(T=50)$ & \meanstd{71.7}{4.29} & \meanstd{0.810}{0.040} & \meanstd{0.717}{0.040}\\
{\ours} $(T=50)$ & \meanstd{74.7}{2.38} & \meanstd{0.826}{0.033} & \meanstd{0.750}{0.024} \\
{\ours} $(T=250)$ & \meanstd{\textbf{82.1}}{1.82} & \meanstd{\textbf{0.976}}{0.014} & \meanstd{\textbf{0.847}}{0.015} \\
\hline
\end{tabular}
}
\label{tab:rebuttal2}
\vspace{-0.15in}
\end{table}

\subsection{Axis Aggregation Scheme Variations}
\label{sec:variation_tokenization}
We varied the number of tokens and the latent dimensionality while keeping the total number of values fixed at $27 \times 3072 = 82{,}944$ and setting $T=256$. In other words, we changed the number of tokens that are combined together in the token aggregation step, where combining a larger number of tokens together leads to a larger token dimensionality while reducing the number of total resulting tokens. Since the tokens are only concatenated and rearranged in this step, the total number of values are kept constant $(82{,}944)$. The results are shown in \cref{tab:rebuttal4}. The results indicate that these adjustments do not lead to significant differences in downstream task performance.
 
\begin{table}[h]
\caption{Results on HCP and ADHD-200 with varying axis aggregation schemes.}
\centering
\resizebox{0.8\linewidth}{!}{
\begin{tabular}{l|cc|cccccc}
\hline
\multirow{3}{*}{Model} & \multirow{3}{*}{\# Token} & \multirow{3}{*}{Dim.} & \multicolumn{3}{c}{HCP}  & \multicolumn{3}{c}{ADHD-200}\\ \cline{4-9} 
& & & \multicolumn{3}{c|}{Intelligence} & \multicolumn{3}{c}{Diagnosis}  \\
 & & & MSE & MAE & \multicolumn{1}{c|}{$\rho$} & ACC & AUC & \multicolumn{1}{c}{F1}\\ \hline
 {\ours} & 27 & 3072 & \meanstd{0.835}{0.053} & \meanstd{0.741}{0.028} & \meanstd{0.392}{0.062} & \meanstd{\textbf{65.8}}{3.50} & \meanstd{0.729}{0.029} & \meanstd{0.630}{0.038} \\
 {\ours\ Alt.1} & 9 & 9216 & \meanstd{\textbf{0.814}}{0.044} & \meanstd{\textbf{0.733}}{0.026} & \meanstd{\textbf{0.416}}{0.061} & \meanstd{65.7}{2.74} & \meanstd{0.710}{0.031} & \meanstd{0.632}{0.043} \\
 {\ours\ Alt.2} & 3 & 27648 & \meanstd{0.840}{0.093} & \meanstd{0.736}{0.044} & \meanstd{0.401}{0.035} & \meanstd{65.1}{3.59} & \meanstd{\textbf{0.690}}{0.032} & \meanstd{\textbf{0.636}}{0.042}  \\
\end{tabular}
}
\label{tab:rebuttal4}
\end{table}

\section{Detailed Experimental Results}
\label{sec:detailed_experimental_results}
We provide the results reported in the manuscript with the standard deviation in Tab. ~\ref{tab:std_3}, Tab.~\ref{tab:std_1}, and Tab.~\ref{tab:std_2}.
\begin{table}[ht!]
\caption{Experimental results with standard deviation on UKB.}
\centering
\resizebox{.85\linewidth}{!}{
\centerline{
\begin{tabular}{l|cccccc}
\hline
\multirow{3}{*}{Method} & \multicolumn{6}{c}{UKB} \\ \cline{2-7}
 & \multicolumn{3}{c|}{Sex} & \multicolumn{3}{c}{Age} \\
 & ACC & AUC & \multicolumn{1}{c|}{F1} & MSE & MAE & \multicolumn{1}{c}{$\rho$} \\ \hline
XGBoost & \meanstd{84.1}{1.7} & \meanstd{0.916}{0.012} & \multicolumn{1}{c|}{\meanstd{0.830}{0.019}} & \meanstd{0.698}{0.013} & \meanstd{0.686}{0.008} & \meanstd{0.553}{0.018}
 \\
BrainNetCNN & \meanstd{91.7}{0.9} & \meanstd{0.969}{0.007} & \multicolumn{1}{c|}{\meanstd{0.912}{0.009}} & \meanstd{0.597}{0.017} & \meanstd{0.618}{0.007} & \meanstd{0.647}{0.012} \\
BNT & \meanstd{92.4}{0.9} & \meanstd{0.980}{0.003} & \multicolumn{1}{c|}{\meanstd{0.919}{0.009}} & \meanstd{0.541}{0.016} & \meanstd{0.588}{0.011} & \meanstd{0.685}{0.011} \\
meanMLP & \meanstd{87.7}{1.8} & \meanstd{0.949}{0.009} & \multicolumn{1}{c|}{\meanstd{0.869}{0.020}} & \meanstd{0.672}{0.031} & \meanstd{0.662}{0.016} & \meanstd{0.586}{0.027}\\
Brain-JEPA & \meanstd{86.8}{0.6} & \meanstd{0.943}{0.004} & \multicolumn{1}{c|}{\meanstd{0.862}{0.007}} & \meanstd{0.688}{0.017} & \meanstd{0.669}{0.008} & \meanstd{0.574}{0.018}\\ \hline
TFF ($T=20$) & \meanstd{98.3}{0.4} & \meanstd{0.998}{0.001} & \multicolumn{1}{c|}{\meanstd{0.982}{0.004}} & \meanstd{0.440}{0.029} & \meanstd{0.525}{0.015} & \meanstd{0.760}{0.015}\\
SwiFT ($T=20$) & \meanstd{97.4}{0.3} & \meanstd{0.998}{0.001} & \multicolumn{1}{c|}{\meanstd{0.972}{0.003}} & \meanstd{0.366}{0.005} & \meanstd{0.480}{0.007} & \meanstd{0.800}{0.004}\\
SwiFT ($T=50$)  & \meanstd{98.1}{0.4} & \meanstd{0.999}{0.001} & \multicolumn{1}{c|}{\meanstd{0.980}{0.005}} & \meanstd{0.364}{0.004} & \meanstd{0.477}{0.005} & \meanstd{0.802}{0.003}\\ \hline
{\ours} ($T=256$) & \meanstd{97.6}{0.2} & \meanstd{0.998}{0.000} & \multicolumn{1}{c|}{\meanstd{0.975}{0.002} }& \meanstd{0.340}{0.011} & \meanstd{0.466}{0.010} & \meanstd{0.814}{0.009}\\ \hline
\end{tabular}
}
}
\label{tab:std_3}
\end{table}

\begin{table}[ht!]
\caption{Experimental results with standard deviation on HCP sex classification and age regression.}
\centering
\resizebox{.85\linewidth}{!}{
\centerline{
\begin{tabular}{l|cccccc}
\hline
\multirow{3}{*}{Method} & \multicolumn{6}{c}{HCP} \\ \cline{2-7}
 & \multicolumn{3}{c|}{Sex} & \multicolumn{3}{c}{Age} \\
 & ACC & AUC & \multicolumn{1}{c|}{F1} & MSE & MAE & \multicolumn{1}{c}{$\rho$} \\ \hline
XGBoost 
& \meanstd{82.2}{2.5} & \meanstd{0.890}{0.028} & \multicolumn{1}{c|}{\meanstd{0.837}{0.025}} & \meanstd{0.859}{0.074} & \meanstd{0.769}{0.033} & \multicolumn{1}{c}{\meanstd{0.296}{0.112}} \\
BrainNetCNN 
& \meanstd{86.3}{4.9} & \meanstd{0.937}{0.027} & \multicolumn{1}{c|}{\meanstd{0.866}{0.049}} & \meanstd{0.847}{0.097} & \meanstd{0.749}{0.040} & \multicolumn{1}{c}{\meanstd{0.372}{0.097}} \\
BNT 
& \meanstd{86.3}{3.0} & \meanstd{0.935}{0.026} & \multicolumn{1}{c|}{\meanstd{0.872}{0.030}} & \meanstd{0.794}{0.051} & \meanstd{0.719}{0.027} & \multicolumn{1}{c}{\meanstd{0.444}{0.055}} \\
meanMLP 
& \meanstd{84.5}{2.5} & \meanstd{0.915}{0.018} & \multicolumn{1}{c|}{\meanstd{0.855}{0.028}} & \meanstd{0.846}{0.056}  & \meanstd{0.751}{0.030}  &   \multicolumn{1}{c}{\meanstd{0.370}{0.087}} \\ 
Brain-JEPA & \meanstd{73.9}{3.2} & \meanstd{0.809}{0.018} & \multicolumn{1}{c|}{\meanstd{0.761}{0.043}} & \meanstd{0.814}{0.037} & \meanstd{0.746}{0.009} &  \meanstd{0.369}{0.046} \\ \hline
TFF ($T=20$) & \meanstd{88.1}{5.0} & \meanstd{0.937}{0.055} & \multicolumn{1}{c|}{\meanstd{0.892}{0.042}} & \meanstd{0.888}{0.062} & \meanstd{0.779}{0.036} & \meanstd{0.246}{0.061}\\
SwiFT ($T=20$) 
& \meanstd{93.1}{0.5} & \meanstd{0.978}{0.008} & \multicolumn{1}{c|}{\meanstd{0.937}{0.004}} & \meanstd{0.776}{0.043} & \meanstd{0.719}{0.015} & \multicolumn{1}{c}{\meanstd{0.450}{0.031}} \\
SwiFT ($T=50$) 
& \meanstd{92.2}{1.1} & \meanstd{0.972}{0.014} & \multicolumn{1}{c|}{\meanstd{0.929}{0.010}} & \textbf{\meanstd{0.764}{0.092}} & \textbf{\meanstd{0.699}{0.047}} & \multicolumn{1}{c}{\meanstd{0.460}{0.071}} \\ \hline
{\ours} ($T=20$) & \meanstd{91.6}{1.5} & \meanstd{0.980}{0.007} & \meanstd{0.923}{0.014} & \meanstd{0.784}{0.120} & \meanstd{0.710}{0.046} & \meanstd{0.460}{0.087} \\
{\ours} ($T=50$) & \meanstd{93.4}{1.3} & \meanstd{0.986}{0.004} & \multicolumn{1}{c|}{\meanstd{0.940}{0.011}} & \meanstd{0.763}{0.069} & \meanstd{0.704}{0.031} & \multicolumn{1}{c}{\meanstd{0.473}{0.051}} \\

{\ours} ($T=256$) & \meanstd{93.8}{0.9} & \meanstd{0.987}{0.003} & \multicolumn{1}{c|}{\meanstd{0.943}{0.008}} & \meanstd{0.773}{0.077} & \meanstd{0.705}{0.038} & \multicolumn{1}{c}{\meanstd{0.473}{0.053}} \\ \hline\hline
{\ours} (3D DCAE) & \meanstd{92.2}{1.7} & \meanstd{0.973}{0.010} & \multicolumn{1}{c|}{\meanstd{0.929}{0.014}} & \meanstd{0.767}{0.118} & \meanstd{0.693}{0.043} & \multicolumn{1}{c}{\meanstd{0.475}{0.076}} \\ 
{\ours} (FT) & \meanstd{95.3}{1.3} & \meanstd{0.986}{0.005} & \multicolumn{1}{c|}{\meanstd{0.958}{0.011}} & \meanstd{0.650}{0.045} & \meanstd{0.655}{0.024} & \meanstd{0.552}{0.032}\\\hline
\end{tabular}
}
}
\label{tab:std_1}
\end{table}
\begin{table}[ht!]
\caption{Main experimental results with standard deviation on HCP intelligence regression and ADHD diagnosis.}
\centering
\resizebox{.85\linewidth}{!}{
\centerline{
\begin{tabular}{l|ccc|ccc}
\hline
\multirow{3}{*}{Method} & \multicolumn{3}{c}{HCP} & \multicolumn{3}{c}{ADHD-200} \\ \cline{2-7}
 & \multicolumn{3}{c|}{Intelligence} & \multicolumn{3}{c}{Diagnosis} \\
 & MSE & MAE & $\rho$ & ACC & AUC & F1 \\ \hline
XGBoost 
& \meanstd{0.908}{0.054} & \meanstd{0.779}{0.023} & \meanstd{0.292}{0.099} & \meanstd{62.3}{2.5} & \meanstd{0.650}{0.036} & \meanstd{0.555}{0.031} \\
BrainNetCNN 
& \meanstd{0.967}{0.119} & \meanstd{0.788}{0.044} & \meanstd{0.286}{0.112} & \meanstd{59.2}{10.7} & \meanstd{0.640}{0.095} & \meanstd{0.545}{0.118} \\
BNT 
& \meanstd{0.920}{0.092} & \meanstd{0.778}{0.054} & \meanstd{0.318}{0.083} & \meanstd{63.6}{5.4} & \meanstd{0.677}{0.062} & \meanstd{0.624}{0.057} \\
meanMLP 
& \meanstd{0.887}{0.076}& \meanstd{0.767}{0.028} & \meanstd{0.340}{0.045} & \meanstd{56.8}{6.8} & \meanstd{0.617}{0.067} & \meanstd{0.532}{0.095} \\ 
Brain-JEPA &\meanstd{0.959}{0.091} &\meanstd{0.799}{0.033} &\multicolumn{1}{c|}{\meanstd{0.171}{0.051}} & -- & -- & -- \\ \hline
TFF ($T=20$) & \meanstd{0.898}{0.022} & \meanstd{0.767}{0.018} & \multicolumn{1}{c|}{\meanstd{0.312}{0.088}} & \meanstd{63.3}{2.3} & \meanstd{0.700}{0.028} & \meanstd{0.608}{0.030}\\
SwiFT ($T=20$) 
& \meanstd{0.940}{0.111} & \meanstd{0.782}{0.044} & \meanstd{0.297}{0.080} & \meanstd{63.3}{3.7} & \meanstd{0.693}{0.030} & \meanstd{0.623}{0.033} \\
SwiFT ($T=50$) 
& \meanstd{0.865}{0.093} & \meanstd{0.758}{0.046} & \meanstd{0.354}{0.070} & \meanstd{63.9}{3.2} & \meanstd{0.701}{0.032} & \meanstd{0.627}{0.030} \\ \hline
{\ours} ($T=20$) & \meanstd{0.866}{0.074} & \meanstd{0.763}{0.042} & \meanstd{0.340}{0.045} & \meanstd{64.4}{3.0} & \meanstd{0.717}{0.020} & \meanstd{0.621}{0.039}\\
{\ours} ($T=50$) & \meanstd{0.916}{0.155} & \meanstd{0.778}{0.063} & \meanstd{0.334}{0.102} & \meanstd{64.3}{2.8} & \meanstd{0.710}{0.021} & \meanstd{0.615}{0.037} \\

{\ours} ($T=256$) & \meanstd{0.835}{0.053} & \meanstd{0.741}{0.028} & \meanstd{0.392}{0.062} & \meanstd{65.8}{3.5} & \meanstd{0.729}{0.029} & \meanstd{0.630}{0.038} \\ \hline\hline
{\ours} (3D DCAE) & \meanstd{0.869}{0.077} & \meanstd{0.755}{0.032} & \meanstd{0.387}{0.078} & \meanstd{65.8}{1.7} & \meanstd{0.711}{0.026} & \meanstd{0.644}{0.022} \\ 
{\ours} (FT) & \meanstd{0.796}{0.051} & \meanstd{0.732}{0.028} & \meanstd{0.435}{0.046} & - & - & - \\\hline

\end{tabular}
}
\label{tab:std_2}
}
\end{table}

\section{Detailed Data Description}
\label{sec:detailed_data_description}
We provide a detailed description of each dataset used in our study in Tab.~\ref{tab:demographics}.

\begin{table}[ht]
  \centering
  \caption{Demographic information of the datasets used in our study}
  \label{tab:demographics}
  \begin{tabular}{l|ccc}
  \hline
    Category            & UKB & HCP              & ADHD
    -200          \\ \hline
    Number of subjects  & 8,178 & 1,061             & 533              \\
    Sex \\
    \hspace{1em}Male, n (\%)       & 4,295 (52.5\%) & 488 (46.0\%)     & 207 (38.8\%)     \\
    \hspace{1em}Female, n (\%)     & 3,883 (47.5\%) & 573 (54.0\%)     & 325 (61.0\%)     \\
    \hspace{1em}N/A, n (\%)         & -- & --               & 1 (0.2\%)        \\
    Age (years) & \meanstd{54.98}{7.53} & \meanstd{28.79}{3.70} & \meanstd{11.94}{3.40} \\
    Intelligence & -- & \meanstd{113.32}{20.50} & -- \\
    Diagnosed, n (\%)  &-- & --               & 236 (44.3\%)     \\ \hline
  \end{tabular}
\end{table}